\pdfoutput=1
\documentclass[sigplan,screen]{acmart}

\usepackage{subcaption}
\usepackage{multirow}
\usepackage{tabularx}
\usepackage{color,soul}
\usepackage{enumitem}
\usepackage{footmisc}
\AtBeginDocument{%
  \providecommand\BibTeX{{%
    \normalfont B\kern-0.5em{\scshape i\kern-0.25em b}\kern-0.8em\TeX}}}

\newcommand{\myhline}{\noalign{\global\arrayrulewidth0.055cm}\hline
                      \noalign{\global\arrayrulewidth1pt}}

\newcommand\appname{\emph{COIN}}
\begin{document}

\title{COIN: \underline{Co}unterfactual \underline{I}mage Generation for VQA Interpretatio\underline{n}}


\author{Zeyd Boukhers \quad Timo Hartmann \quad Jan Jürjens}
\email{{boukhers,tihartmann, juerjens}@uni-koblenz.de}
\affiliation{%
  \institution{\textit{
  University of Koblenz-Landau}}
  \city{Koblenz}
  \country{Germany}
}



\begin{abstract}
Due to the significant advancement of Natural Language Processing and Computer Vision-based models, Visual Question Answering (VQA) systems are becoming more intelligent and advanced. However, they are still error-prone when dealing with relatively complex questions. Therefore, it is important to understand the behaviour of the VQA models before adopting their results. In this paper, we introduce an interpretability approach for VQA models by generating counterfactual images. Specifically, the generated image is supposed to have the minimal possible change to the original image and leads the VQA model to give a different answer. In addition, our approach ensures that the generated image is realistic. Since quantitative metrics cannot be employed to evaluate the interpretability of the model, we carried out a user study to assess different aspects of our approach. In addition to interpreting the result of VQA models on single images, the obtained results and the discussion provides an extensive explanation of VQA models' behaviour.

\end{abstract}



\maketitle

\section{Introduction}
\label{sec:intro}

Over the past years, the task of Visual Question Answering (VQA) has been widely investigated taking advantage of the development strides of Natural Language Processing and Computer Vision. A VQA model aims to answer a natural language question about the content of an image or one of the appearing objects. Due to the complexity of the task, VQA models are still in the early stage of research and up to our knowledge, is not integrated into any running system. One of the inherent problems of VQA systems is the reliance on the correlation between the question and answer more than the content of the image~\cite{chen2020counterfactual,niu2021counterfactual}. Furthermore, the available datasets are usually unbalanced regarding certain types of questions. In the VQAv1 dataset \cite{antol2015vqa}, for instance, simply answering tennis to any sports-related question without considering the image yields an accuracy of approximately 40\%. This is because the dataset creators tend to generate questions about objects detectable in the image which make the dataset suffer from the so-called ``\emph{visual priming bias}''. For example, blindly answering ``\emph{yes}'' to all questions starting with ``\emph{Do you see a...?}'' without considering anything else yields approximately 90\% accuracy in the VQAv1 dataset \cite{niu2021counterfactual, goyal2017making}. In practice, this bias is not distinctly perceivable because the users tend to ask similar questions related to the image of the appearing objects and they most likely know the correct answer. For more complex questions or when the user lacks the knowledge expertise of the questions or the image content (e.g. medical domain), it won't be possible to capture the behaviour of the VQA system and whether it is biased. Therefore, it is important to interpret the result of these systems and find what caused the model to output a given answer based on the image-question pair.


Although there is no uniform definition of ``\emph{Interpretability}'', researchers agree that the interpretability of ML models increases the users' trust in ML systems. For VQA models, there exist a limited number of papers that investigate the interpretation of their models. Existing attempts include (i) the identification of visual attributes that are relevant to the question \cite{zhang2019interpretable, li2018tell} and (ii) the generation of counterfactuals  \cite{niu2021counterfactual, pan2019question, teney2020learning}.

In the direction of (i), the method proposed by Zhang et al. \cite{zhang2019interpretable} generates a heatmap over the input image to highlight the image regions that are relevant to the answer of the VQA model. However, as pointed out by Fernández-Loría et al. \cite{fernandez2020explaining}, this approach explains the model's prediction but does not provide a sufficient explanation of its decision. They suggest instead that counterfactual explanations offer a more sophisticated way to increase the interpretability of an ML model because they reveal the causal relationship between features in the input and the model’s decision. For example, considering an ML model that classifies MRI images into \emph{Malignant} or \emph{Benign}, the generated heatmap can highlight the key ROI of the model's prediction. Although this heatmap answers the question \emph{What did lead to this decision?}, it does not answer an important question \emph{Why did it lead to this decision?} Consequently, this approach does not provide insight into how the model would behave under alternative conditions.  To provide more in-depth interpretability, generating a counterfactual image that is minimally different from the original one but leads to a different model's output would indirectly answer the question \emph{Why did the model take such a decision?} To the best of our knowledge, only three existing methods \cite{chen2020counterfactual, pan2019question, teney2020learning} aim at making VQA models interpretable by providing counterfactual images.

Chen et al. \cite{chen2020counterfactual} introduce a method that generates counterfactual samples by applying masks to critical objects in images or words in questions. Similarly, Teney et al. \cite{teney2020learning} suggest a method that masks features in images whose bounding boxes overlap with human-annotated attention maps. Finally, in their ongoing research, Pan et al. \cite{pan2019question} propose a framework to generate counterfactual images by editing the original image such that the VQA model returns an alternative answer for a given question. Due to the complexity of the problem, the approach is restricted to colour-based questions. Given a tuple (Image, Question and the VQA's answer), the approach first finds the question-critical object and then changes its colour so that the VQA system gives a different answer. However, this change is not limited to the question-critical object but all regions with similar colours are changed. Given that the main goal of interpreting VQA systems is to help an expert user understand the behaviour of the VQA model, the approach presented in \cite{pan2019question} requires that the user understands the relationship between the image, the question and the answer.

As the user needs interpretation mainly when he lacks the necessary knowledge to understand the relationship between the input and output, we propose to extend the work of Pan et al. \cite{pan2019question} by restricting the change to the question-critical object. To this end, this paper introduces an attention mechanism that identifies question-critical objects in the image and guides the counterfactual generator to apply the changes on specific regions. Moreover, a weighted reconstruction loss is introduced in order to allow the counterfactual generator to make more significant changes to question-critical spatial regions than the rest of the image. For further improvement and future work, we made the entire implementation of the guided generator publicly available. In summary, this paper aims to best interpret the output of VQA models by answering the following research questions: 

\begin{itemize}[leftmargin=*]
    \item \textbf{RQ1:} How to change the answer of a VQA model with the minimum possible edit on the input image?
    \item \textbf{RQ2:} How to alter exclusively the region in the image on which the VQA model focuses to derive an answer to a certain question? 
    \item \textbf{RQ3:} How to generate realistic counterfactual images? 
    
\end{itemize}


Following this section, Section~\ref{sec:re_wo} discusses the related works. Section~\ref{sec:method} presents the proposed approach and Section~\ref{sec:exp} presents the conducted experiments and the obtained results that validate the effectiveness of the proposed approach. Finally, Section~\ref{sec:conc} concludes this paper and gives insight into future directions.

\section{Related Work}
\label{sec:re_wo}

This papers addresses the problem of interpreting the outcome of VQA systems. Therefore, we will review in this section the related works divided into three categories: (1) Interpretable Machine Learning, (2) Visual Question Answering (VQA) and (3) Interpretable VQA. 

\subsection{Interpretable Machine Learning}
Throughout the past decades, the notion of interpretability increasingly gained attention in the ML domain \cite{chakraborty2017interpretability, doshi2017towards, gilpin2018explaining, kim2016examples, shwartz2017opening, tsang2018neural, zhang2018visual, zhang2020interpretable}. According to Kim et al. \cite{kim2016examples}, interpretability is particularly important with respect to systems whose decisions have a significant impact such as healthcare, criminal justice systems and finance. 
Interpretability serves several purposes, including protecting certain groups from being discriminated against, understanding the effect of parameter and input variation on the model's robustness and increasing the user's trust in automated intelligent systems \cite{doshi2017towards}. Therefore, a model is considered interpretable if it allows a human to consistently and correctly classify its outputs \cite{kim2016examples} and understand the reason behind the model's output \cite{chakraborty2017interpretability}. ML models such as decision trees are inherently interpretable, as they provide explanations during training or while the output is generated. However, most of the sophisticated ML models used nowadays such as deep neural networks are not interpretable by nature and require post-hoc explanations \cite{moraffah2020causal}


One way to create such explanations is to use global surrogate models, where the aim is to approximate the prediction function $f$ of a black-box and complex model (e.g. neural networks) to the best possible with a prediction function $g$ such that $g$ is the prediction function of an inherently interpretable model (e.g. decision trees or linear regression) \cite{nobrega2019towards, moraffah2020causal}. Another way is to use local surrogate models, which individually explain the predictions of a trained ML model to have an overview of its behaviour \cite{nobrega2019towards}. Ribeiro et al. \cite{ribeiro2016should}  propose a Local Interpretable Model-agnostic Explanations (LIME) which approximates the output of black-box models by examining how variations in the training data affect its predictions. Particularly, LIME permutes a trained black-box model’s training samples to generate a new dataset. Based on the black-box model’s predictions on the permuted dataset, LIME trains an interpretable model, which is weighted by the proximity of the sampled instances to the instance of interest.

Feature visualization is another direction to increase the interpretability of black-box models. The goal is to visualize the features that maximize the activation of a NN's unit \cite{nobrega2019towards}. This direction takes advantage of the structure of NNs, where the relevance is backpropagated from the output layer to the input layer \cite{zhang2018visual}. Mainly, most of the approaches under this direction are dedicated to image classification tasks by providing a saliency map that highlights the pixels relevant to the model's output \cite{montavon2017explaining, moraffah2020causal, simonyan2013deep}. A common interpretability direction suggests generating example-based explanations for complex data distributions. The aim is to find prototypes from the training dataset that summarize the prediction of the model \cite{moraffah2020causal}. Although this approach can satisfy the user need for interpretation in simple tasks, it is not practical for most of the real-world data which are heavily complex and seldom contain representative prototypes \cite{kim2016examples}. Therefore, Kim et al. \cite{kim2016examples} propose to identify some criticism samples that deliver insights about those prototypes which is not covered by the model.

The problem of post-hoc interpretability methods is their incapability to answer how the model would behave under alternative conditions (e.g. different training data). Therefore, causal interpretability approaches aim at finding why did the model make a certain decision instead of another one or what would be the output of the model for a slightly different input \cite{moraffah2020causal}. Here, the goal is to extract causal relationships from the data by analysing whether changing one variable cause an effect in another one \cite{murdoch2019interpretable}. One way to achieve this is by finding a counterfactual input that affects the model's prediction \cite{denton2019image, gomez2020vice, goyal2019counterfactual, hendricks2018generating, sokol2018glass, verma2020counterfactual}. For instance, Goyal et al. \cite{goyal2019counterfactual} propose a method that identifies how a given image ``$I$'' could change so that the image classifier outputs a different class by replacing the key discriminative regions in ``$I$' with pixels from an identified "distractor" image ``$I_0$' that has a different class label. Pearl \cite{pearl2018theoretical} suggests that generating counterfactuals allows for the highest degree of interpretability among all methods to explain black-box models.

\subsection{Visual Question Answering (VQA)}
Taking advantage of the remarkable advancement of computer vision and natural language processing, due to Deep Neural Networks, several research works have addressed the task of VQA in the past years \cite{chen2020counterfactual, goyal2017making, antol2015vqa, malinowski2014multi, srivastava2019visual, zhang2016yin, anderson2018bottom, geman2015visual, gupta2017survey}. According to Antol et al. \cite{antol2015vqa}, VQA methods aim at answering natural language questions about an input image. This combination of image and textual data makes VQA a challenging multi-modal task that involves understanding both the question and the image \cite{srivastava2019visual}. The answer’s format can be of several types: a word, a phrase, a binary answer, a multiple-choice answer, or a ``fill in the blank'' answer \cite{srivastava2019visual, gupta2017survey}.


In contrast to earlier contributions, recent VQA approaches aim at generating answers to free-form open-ended questions \cite{wu2017visual}. Agrawal et al. \cite{antol2015vqa}, for example, propose a system that classifies an answer to a given question about an image by combining a CNN-based model to extract features from the image and Long Short-Term Memory (LSTM)-based model to process the question. This model, referred to as \emph{Vanilla VQA}, can be considered as a benchmark for DL-based VQA methods \cite{srivastava2019visual}. Yang et al. \cite{yang2016stacked} introduce \emph{Stacked Attention Networks (SANs)} that uses CNNs and LSTMs to compute an images’ regions related to the answer based on the semantic representation of a natural language question. Similarly, Anderson et al. \cite{anderson2018bottom} propose to narrow down the features in images by using top-down signals based on a natural language question to determine what to look for. These signals are combined with bottom-up signals stemming from a purely visual feed-forward attention mechanism.

Despite the continuous advancements in VQA, several papers suggest that VQA models tend to suffer from the language prior problem, where the models tend to achieve good superficial performances but do not truly understand the visual context \cite{chen2020counterfactual, goyal2017making, zhang2016yin, zhu2020overcoming}. Specifically, Goyal et al. \cite{goyal2017making} found that in the VQAv1 dataset \cite{antol2015vqa} blindly answering "\emph{yes}" to any question starting with "\emph{Do you see a...?}" without taking into account the rest of the question or the image results in an accuracy of 87\%. To overcome this, they proposed a balanced dataset to counter language biases, such that for a given triplet (image $I$; question $Q$; answer $A$) from the VQAv1 dataset \cite{antol2015vqa}, humans were asked to identify a similar image $I’$ for which the answer to question $Q$ is different from $A$. Similarly, Zhang et al. \cite{zhang2016yin} propose a balanced VQA dataset for binary questions, where for each question, pairs of images showing abstract scenes were collected so that the answer to the question is "\emph{yes}" for one image and "\emph{no}" for the other.

Moreover, Chen et al. \cite{chen2020counterfactual} assume that existing VQA models capture superficial linguistic correlations between questions and answers in the training set and, hence, yields low generalizability. Therefore, they propose a model-agnostic Counterfactual Samples Synthesizing (CSS) training scheme that aims at improving VQA models’ visual-explainable and question-sensitive abilities. The CSS algorithm masks (i) objects relevant to answering a question in the original image to generate a counterfactual image and (ii) critical words to synthesize a counterfactual question. In the same context, Zhu et al. \cite{zhu2020overcoming} propose a self-supervised learning framework that balances the training data but first, identifies whether a given question-image pair is relevant (i.e., the image contains critical information for answering the question) or irrelevant. This information is then fed to the VQA model to overcome language priors.

These above-discussed problems indicate that the empirical results of VQA models do not reflect their efficacy, especially when promoting VQA models to serve their intended purposes. Specifically, answering questions that the user cannot answer, such as in the healthcare domain. Therefore, it is important to make the output of the VQA model interpretable and not only rely on the evaluation results.

\subsection{Interpretable VQA}

In most real-world scenarios, human users want to get an explanation along with a VQA system’s output, especially if it fails to answer a question correctly or when the user does not know whether the answer is correct \cite{li2018tell}. However, there exist only a few papers addressing the task of interpreting and explaining the outcome of VQA models \cite{niu2021counterfactual, das2017human, li2018tell, pan2019question, zhang2019interpretable}. Also, most of the existing approaches rarely provide human-understandable explanations regarding the mechanism that led to a given answer. Li et al. \cite{li2018tell} introduce a method that simulates the human question-answering behaviour. First, they apply pre-trained attribute detectors and image captioning to extract attributes and generate descriptions for the given image. Second, the generated explanations are used instead of the image data to infer an answer to a question. Consequently, providing critical attributes and captions to the end-user allows them to understand better what the system extracts from the image.  Zhang et al. \cite{zhang2019interpretable} introduce a heat map-based system to display the image’s regions relevant to the question to the user. To this end, they employed in their model region descriptions and object annotations provided in the Visual Genome dataset \cite{krishna2016visual}.

Pan et al. \cite{pan2019question} introduce a method that provides counterfactual images along with a VQA model’s output. Precisely, for a given question-image pair, the
system generates a counterfactual image that is minimally different from the original image and visually realistic but leads the VQA model to output a different answer for the given question. In its current form, their method is restricted to the context of colour questions. Furthermore, since their model makes edits on a pixel-by-pixel level, the counterfactual images contain changes also in areas irrelevant to a given question. Therefore, the counterfactual image does not provide any meaningful interpretability if the question-critical object has the same relative colour as other objects.

\section{Method}
\label{sec:method}

In this paper, we propose a method named \emph{COIN} that provides human-interpretable discriminatory explanations to increase a VQA system’s interpretability. The aim is to interpret the outcome of the VQA model by answering the question: ``\emph{How would the image look like so that the VQA model gives a different outcome?}''. Concretely, given an image-question pair $(I; Q)$ and a VQA model $f : (I, Q) \rightarrow A$, where $A$ is its predicted answer, the goal is to train a model $\mathcal{G}$ to generate a new image $I'$, such that:

\begin{equation}
    \mathcal{G}: (I, Q, A) \rightarrow I' \quad | \quad f : (I', Q) \rightarrow A' \quad ;  \quad  \forall A' \neq A,
\end{equation}

\noindent where $I'$ is the counterfactual image of $I$. Since there are infinite possible images that can satisfy the above constraint, along the lines of Pan et al. \cite{pan2019question}, \appname~ aims to tackle the research question \emph{\textbf{RQ1}} by constraining $\mathcal{G}$ to (i) be different as minimally as possible from $I$, (ii) be visually realistic, (iii) contain semantically meaningful edits and (iv) be applied only on the question-relevant regions. Intuitively, with these constraints, we aim to change the output of the VQA model by applying as minimum as possible changes only on the semantically relevant object so that the user can perceive what can change the output of the VQA model. To this end, we propose to extend the counterfactual GAN introduced by Pan et al. \cite{pan2019question} by allowing for semantic edits beyond colour modifications (e.g., editing shapes) and ensuring that only the question-critical regions in an image are altered while retaining the background. The  variables  used  for  the system definition are summarized in Table \ref{tab:app:var}.

\begin{table}[]
    \centering
    \begin{tabularx}{\linewidth}{l|X}
    \myhline
    \textbf{Variable} & \textbf{Description} \\
    \hline
        $\mathcal{G}$   &  The counterfactual generator proposed in this paper.\\
   $f$  &  The VQA model (i.e. MUTAN~\cite{ben2017mutan} in this paper)\\
   \hline
   $I$ & Original image \\
   $Q$ & Question about $I$\\
   $A$ & The answer of $f$ to $Q$ given $I$\\
   \hline
   $h$ & $I$'s height\\
   $w$ & $I$'s width \\
   \hline
   $I'$ & The counterfactual image of $I$, generated by $\mathcal{G}$\\
   $A'$ & The answer of $f$ to $Q$ given $I'$\\
   $\hat{I}$ & An image generated by $\mathcal{G}$\\
   \hline
   $M$ & The attention map of $I$ \\
   $M'$ & The attention map of $I'$ \\
   \myhline
    \end{tabularx}
    \caption{Summary  of  variables  used in this paper}
    \label{tab:app:var}
\end{table}

Figure.~\ref{fig:app:overview} illustrates an overview of the proposed architecture. In the depicted example, given the image $I$, the answer of the VQA model to the question ``\emph{$Q$: What is the colour of the bird?}'' is ``\emph{$A$: Blue and White?}''. To explain this output,  $\mathcal{G}$ has to abide by the above-mentioned constraints. To this end, $\mathcal{G}$ consists of several components:


\begin{figure*}[h]
    \centering
    \includegraphics[width=\linewidth]{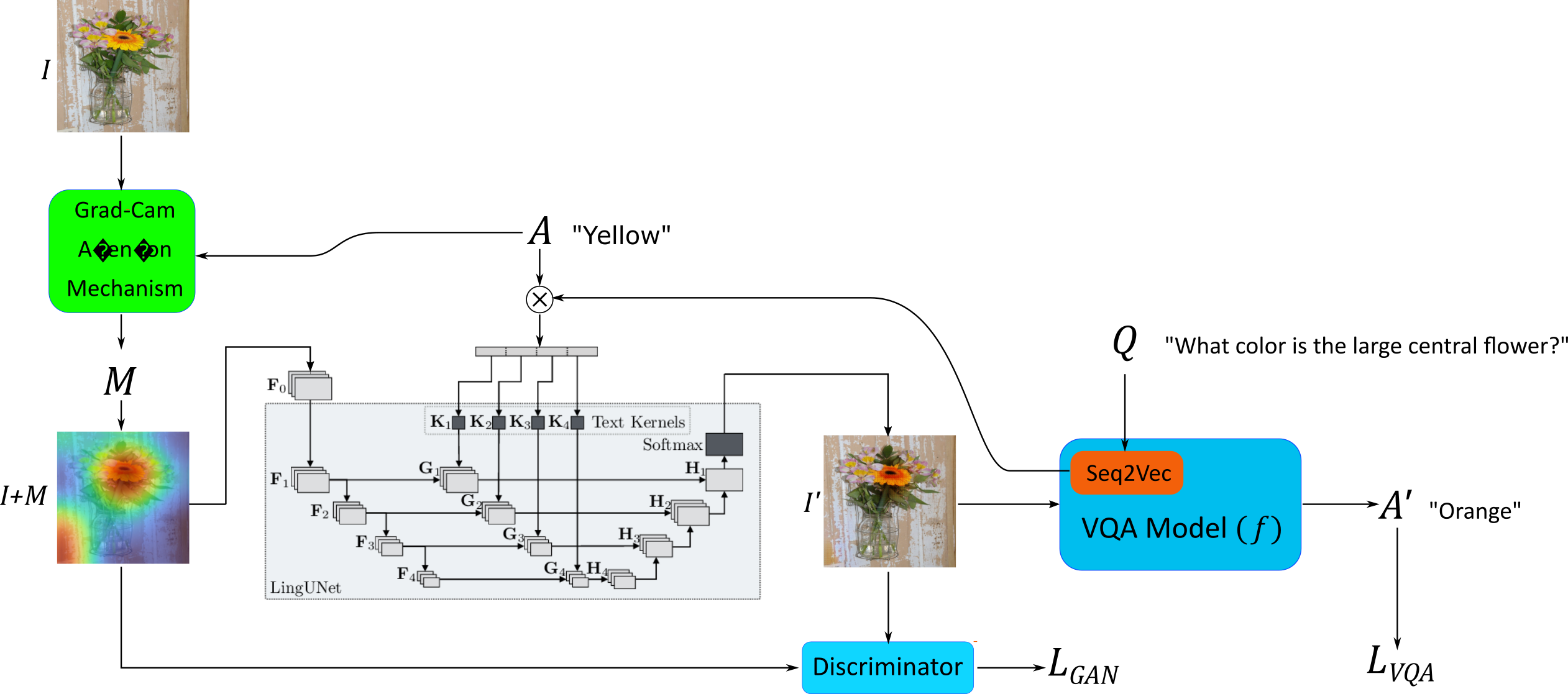}
    \caption{Overview of the proposed architecture inspired by Pan et al. \cite{pan2019question}}
    \label{fig:app:overview}
\end{figure*}

\subsection{ROI Guide}

\begin{figure}
    \centering
    \includegraphics[width = \linewidth]{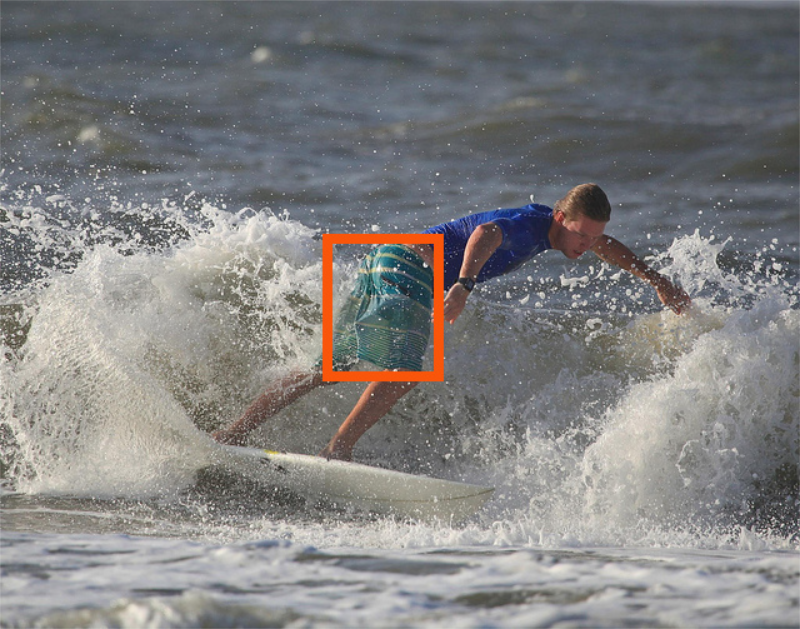}
    \caption{Example question-image pair from the VQAv1 dataset \cite{antol2015vqa}. The red bounding box indicates the question-critical object.}
    \label{fig:app:attention}
\end{figure}

To tackle the research question \emph{\textbf{RQ2}}, $\mathcal{G}$ has to be guided  to primarily edit regions in $I$ that are relevant to $Q$. To this end, \appname~ aims to identify the question-critical object in $I$, but, complex images may contain various objects, of which, usually, only one or a few are relevant when answering a given question. Therefore, an object can be considered to be question-critical if it is key to finding an answer to a given question. For example, given the question ``\emph{What colour are the man’s shorts?}'', the question-critical object in Figure.~\ref{fig:app:attention} is the man’s shorts. Therefore, \appname~ aims to guide the generator with a continuous attention map $ M\in \mathbb{R}^{1 \times h\times w}$ in the range $[0,1]$, where $h$ and $w$ correspond to the input image’s height and width, respectively. This map is supposed to highlight the discriminative ROIs of the image $I$ that led $f(I)$ to output the answer $A$. Thus, instead of generating a counterfactual image $I'$ based on the original image $I$, the latter is concatenated with the attention map $M$, such that the concatenation $[I; M]$ serves as an input to the generator $\mathcal{G}$.

To obtain $M$, an attention mechanism is used to determine each pixel’s importance regarding the VQA model’s decision. The intuition is to identify the spatial regions in an image that are most relevant to answer a given question. For this reason, \appname~ applies the Gradient-weighted Class Activation Mapping (Grad-CAM) algorithm \cite{selvaraju2017grad} to the VQA model’s final convolution layer because convolution layers retain spatial information that is not kept by fully connected layers. Hence, CNN’s last convolution layer is supposed to have the finest balance between high-level semantics and fine-grained spatial information. Grad-CAM exploits this property by finding the gradient of the most dominant logit (i.e., in the case of a VQA model, this corresponds to the answer with the highest probability) that flows into the model’s final activation map. Furthermore, since Grad-CAM is suitable for various CNN-based models, it can be applied to most VQA models.

Intuitively, the algorithm computes the importance of each neuron activated in the CNN’s final convolutional layer with respect to its prediction. Computing the gradient $y^a$ of the logit corresponding to the VQA model’s predicted answer $a$ with respect to the $k$th feature map’s activations $\phi^k$ of a convolutional layer, i.e. $\frac{\delta y^a}{\delta \phi^k}$, reveals the localization map $L^c_{Grad-CAM} \in \mathbb{R}^{u\times v}$ of width $u$ and height $v$. Next, channel-wise pooling with respect to the width and height dimensions is applied to the gradients. The pooled gradients are then used to weigh the activation channels. Finally, the weighted activations $\alpha^a_k$ reveal each channel’s importance with respect to the VQA model’s prediction \cite{selvaraju2017grad}:

\begin{equation}
    \alpha^a_k=\frac{1}{Z} \sum_i^u{\sum_j^v{\frac{\delta y^a}{\delta \phi^k_{i,j}}}}
\end{equation}

Performing a weighted combination of forward activation maps followed by a ReLU finally yields a coarse saliency map of the same size as the convolutional feature maps \cite{selvaraju2017grad}:

\begin{equation}
    L^c_{Grad-CAM}=ReLU\left( \sum_k{\alpha^a_k \phi^k} \right)
\end{equation}

Finally, to obtain $M$, the feature maps $L^c_{Grad-CAM}$ are interpolated to match the size of the input image $I$. Furthermore, a gaussian filter with a mean $\mu = 0$ and a population standard deviation $\sigma = 2$ is applied for improved preservation of the selected image regions’ edges \cite{bergholm1987edge}.



\subsection{Language-Conditioned Counterfactual Image Generation}

To drive $\mathcal{G}$ generates a counterfactual image $I'$ such that the corresponding answer $A' \neq A$, \appname~ follows Pan et al. \cite{pan2019question} by adopting an architecture based on LingUNet \cite{misra2018mapping}, which is an encoder-decoder NN similar to the popular pixel-to-pixel UNet model \cite{ronneberger2015u} and maps conditioning language to key intermediate filter weights based on an embedding of natural language text.

Similarly to \cite{pan2019question}, \appname~ feed $\mathcal{G}$ with language embedding which is the concatenation of the VQA model’s question encoding and answer encoding. The question encoding is represented by the question embedding $\overline{q}$, which stems from the VQA model's language encoding for the question $Q$. The answer encoding is represented by the answer embedding $\overline{a}$, which is the VQA's final logits weight vector w.r.t its prediction $A$ for the image-question pair $(I, Q)$. The goal here is to train $\mathcal{G}$ with the VQA model’s negated cross-entropy for $A$ being the target. Consequently, the generated image $I'$ should contain semantically meaningful differences compared to $I$, such that the VQA model outputs two different answers for $I'$ and $I$ given the same question $Q$. 

Precisely, $\mathcal{G}$ applies a series of operations to condition the image generation process on language. First, the question embedding $\overline{q}$ and the answer embedding $\overline{a}$ are concatenated to create a language representation $\overline{x}$. Second, $\mathcal{G}$ applies a 2D $1 \times 1$ convolutional filter with weights $K_k$ to each feature map $F_j$ . Each $K_k$ is computed by splitting $\overline{x}$ into $m$ equally sized vectors $\{\overline{x}\}_{j=1}^{m}$ and applying a $1 \times 1$ linear transformation to each of them. Applying the filter weights to each $F_j$ yields the language-conditioned feature maps $G_j$ \cite{misra2018mapping}.

Next, LingUNet performs a series of convolution and deconvolution operations to generate a new image $\hat{I}$. The final counterfactual $I'$ is retrieved as follows:

\begin{equation}
    I'=M \odot \hat{I} + (\mathbf{1}-M) \odot I, 
\end{equation}

\noindent where $\odot$ denotes the element-wise multiplication and $\mathbf{1}$ is an all-ones matrix with the same dimension. Intuitively, $I'$ is created by incorporating to the original image’s background, the foreground of $\hat{I}$, which is denoted by pixels with large attention values representing a higher intensity compared to those with low attention values.

\subsection{Minimum change} 
 Although $I$ and $I'$ should have distinct semantics with respect to a given question Q, the differences between the two images should be as minimal as possible. To this end, \appname~ incorporates a reconstruction loss, which penalizes the generator for creating outputs different from the input. To ensure that question-critical objects can change their semantic meaning, the generator should be allowed to make significantly more changes in the corresponding image regions (i.e. the foreground) than in the rest of the image (i.e., the background). A modified $\ell_2$-loss adapted to this purpose, which incorporates the attention map $M$ as a relative weighting term, acts as the reconstruction loss:

\begin{equation}
    \ell_2=\left[ || \left(\mathbf{1}-M\right) \odot I- \left(\mathbf{1}-M\right) ||_2^2\right].
\end{equation}
    
Applying a weighted reconstruction loss aims at contributing to the desired traits that (i) the model predominantly edits critical objects and (ii) a relatively $\ell_2$-loss constraint is applied to question-critical regions, allowing for more significant semantic edits. Contrarily, the stricter $\ell_2$- constraint for question-irrelevant regions ensures that the generator retains them nearly unchanged.

\subsection{Realism}

The counterfactual images generated by $\mathcal{G}$ should be visually realistic. To this end and to tackle the research question (\emph{\textbf{RQ3}}), \appname~ employs a PatchGAN discriminator as proposed by Isola et al. \cite{isola2017image}. This discriminator learns to distinguish between real and fake images and penalizes unrealistic generated counterfactual images. The generator and the discriminator are trained in an adversarial manner as in GAN training \cite{goodfellow2020generative}.

\noindent \paragraph{Spectral Normalization for stabilize training:}
Training GANs can suffer from instability and be vulnerable to the problems of exploding and vanishing gradients \cite{lin2020spectral}. In their approach, Pan et al. \cite{pan2019question} applied gradient clipping to counter this problem, which requires extensive empirical fine-tuning of the training regime. To bypass this extensive procedure,
\appname~ uses spectral normalization \cite{lin2020spectral, miyato2018spectral} to counteract training instability as Miyato et al. \cite{miyato2018spectral} suggest that using spectral normalization in GANs can lead to the generated images having a higher quality relative to other training stabilization techniques, such as gradient clipping. Given a real function $\gamma: \mathbb{R} \rightarrow \mathbb{R}$, the Lipschitz constraint is followed if $|\gamma(x_i)-\gamma(x_2)|/|x1-x2| \leqslant k$, where $k$ is the Lipschitz constanz (e.g., $k=1$).  Given a CNN $\mathcal{CNN_{\theta}}$ with $L$ layers and weights $\theta = \{ w_1, w_2,\cdots, w_L\}$, its output for an input $x$ can be computed as \cite{lin2020spectral}:

\begin{equation}
    \mathcal{CNN_{\theta}}= a_L \odot l_{w_L} \odot a_{L-1} \odot l_{w_{L-1}}  \odot \cdots \odot  a_1 \odot l_{w_1} (x),
\end{equation}

\noindent where $a_{i=1}^{L}$ denotes the activation function in the $i$th layer. Spectral normalization regularizes the convolutional kernels $w_i \in \mathbb{R}^{c_{out}\times c_{in} \times k_w \times k_h}$ with kernel width $k_w$ and height $k_h$ of the fully connected layers $l_{w_i}$ and $c_{in}$ and $c_{out}$ be the input and output channels, respectively. To this end, $w_i$ is first reshaped into a matrix $\hat{w}_i \in \mathbb{R}^{c_{out}} \times ( c_{in} \times k_w \times k_h)$, which is then normalized such that the spectral norm $||\hat{w}_i||_{sp} = 1 \forall i=1,\cdots,L$. Thereby, the spectral norm is computed as follows \cite{lin2020spectral}:

\begin{equation}
    ||\hat{w}_i||_{sp}=\frac{\hat{w}_i}{u_i^T \times \hat{w}_i \times v_i},
\end{equation}

\noindent where $u_i$ and $v_i$ denote the left and sight singular vectors of $\hat{w}_i$ with respect to its largest singular value. 

\section{Experiments}
\label{sec:exp}

In this section, we evaluate the effectiveness of the proposed approach from different aspects, namely, the capability of $\mathcal{G}$ to (i) generate counterfactual image $I'$ such that $f(I';Q) \neq f(I;Q)$ (\emph{\textbf{RQ1}}), (ii) focus the changes on the object-critical object (\emph{\textbf{RQ2}}), (iii) generate realistic images (\emph{\textbf{RQ3}}). For result reproducibility and further improvements, we made our code and results publicly available under this link~\footnote{\url{https://coin.ai-research.net/} \label{fn:url}}

\subsection{Dataset}

For all our experiments, we used a subset of the VQAv1 dataset’s \emph{Real Images} portion introduced by Agrawal et al. [5] The dataset covers (i) images of everyday scenes with a wide variety of questions about the images and the corresponding ground truth answers. Despite the dataset includes samples with several types of questions, for feasibility reasons, we focus in this experiment on color- and shape-based questions only. This yields a set of 23,469 tuples (Image, Question, Answer). The subset is publicly available under this link~\footref{fn:url}

\subsection{VQA model}
As a VQA model, we employed MUTAN \cite{ben2017mutan}, which is trained on VQAv1 dataset. MUTAN achieved overall accuracy of approximately $67\%$ and it performed particularly well on questions with binary \emph{Yes/No} answers (accuracy $\approx 85.14\%$). For quantitative questions (e.g., ``\emph{How many
...?}''), it achieved and accuracy of $39.81\%$. For all other question types, including color and shape-based questions, it achieved an accuracy of $58.52\%$.

\subsection{Evaluation and Results}

Automatically and objectively assessing the quality of synthetically generated images is a challenging task \cite{nam2018text, salimans2016improved, zhu2021image}. Salimans et al. \cite{salimans2016improved} suggest that there exists no objective function to assess a GAN’s performance. Furthermore, the goal of \appname~ is to provide an understandable interpretability to the VQA output. The quality of this interpretability can only be assessed by the satisfaction of the user. Therefore, we conducted a user study by presenting the output of \appname~ (i.e. $I'$ and $A')$ together with $I$,  $A$ and $Q$. For every sample, the user answers five questions divided in two phases: 

\begin{itemize}[leftmargin=*]
    \item \textbf{Phase I:} We present the participant with $I'$, $Q$ and  $A'$. The participant is requested then to answer the following questions: 
    \begin{enumerate}
        \item \emph{Is the answer correct?}\\
            with three possible answers: \emph{Yes}, \emph{No}, and  \emph{I am not sure}. 
        \item \emph{Does the picture look photoshopped: any noticeable edit or distortion (automatic or manual)?}
           with five possible answers (i.e. from Very real to Clearly photoshopped).
    \end{enumerate}
    
    \item \textbf{Phase II:} We present the participant with $Q$ and both images $I$ and $I'$ together with the answers of the VQA model $A$ and $A'$. The participant is requested then to answer the following questions: 
    \begin{enumerate}
        \item \emph{Which of the images is the original?}
            with three possible answers: \emph{Image 1}\footnote{Note that we do not give any information about the image. \label{fn:image}}, \emph{Image 2}\footref{fn:image} and \emph{I am not sure}
        \item \emph{Is the difference between both images related to the question-critical object?}
           with three possible answers: \emph{Yes}, \emph{No} and \emph{I am not sure}
        \item \emph{Which pair (Image, Answer) is correct?}
           with four possible answers: \emph{Image 1}\footref{fn:image}, \emph{Image 2}\footref{fn:image}, \emph{Both} and \emph{None}
    \end{enumerate}
\end{itemize}

To make this experience easy for the participants, we built a web application~\footref{fn:url}, where $94$ participants have participated in the survey answering the above questions for $1320$ unique samples. Note that some samples have been treated by more than one participant, but maximum by maximum three, which make the total number of samples $2001$. In the following, we present the qualitative and quantitative (obtained from the user study) results of \appname~ w.r.t the above-mentioned evaluation aspects:

\paragraph{Semantic change (RQ1):}
Since the final goal of \appname~ is to generate an counterfactual image that drives the VQA model to give a different answer, we evaluate here the capability of $\mathcal{G}$ to generate these images. Among $12096$ counterfactual images generated by $\mathcal{G}$, $37.82\%$ of them lead $f$ to output an answer $A' \neq A$. In particular, $f$ outputs a different answer for the counterfactual than for the original image for $38.05\%$ of the color-based questions and $25.45\%$ of the shape-based questions. Among the samples treated in our survey, the participants found that the VQA outputs a correct answer $A$ given $I$ and $Q$ for only $45.8\%$ of the presented images, while it outputs a wrong answer for $41.4\%$ of them. In the remaining $12.8\%$ of the images, the participants couldn't decide because (i) the question was not understood, (ii) the correct answer is not unique or (iii) the answer is only partially correct. After presenting the participants with both images $I$ and $I'$, the question $Q$ and the answers $A$ and $A'$, the participants changed their opinions about $484$ samples, where they found that $A$ is correct for $70.5\%$ of the presented images. This indicates that the participants could understand the question and answer better after interpreting the result of $f$. For the generated images, the participants found that $f$ outputs a correct answer $A'$ for only $40.5\%$.

\begin{figure*}
\centering
    \begin{subfigure}[b]{0.8\textwidth}
    \vspace{-0.5cm}
    \caption{\textbf{Question}: What color are the peppers in the bottom left corner?}
     \begin{subfigure}[b]{0.5\textwidth}
     \includegraphics[width=0.48\textwidth]{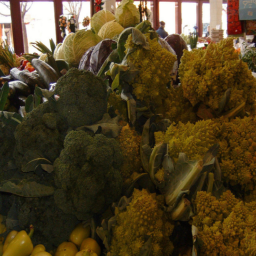}
     \includegraphics[width=0.48\textwidth]{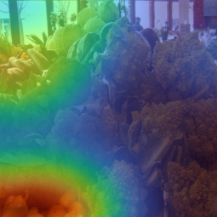}
     \caption*{\textbf{$f$'s answer}: yellow}
     \end{subfigure}
     \begin{subfigure}[b]{0.5\textwidth}
     \includegraphics[width=0.48\textwidth]{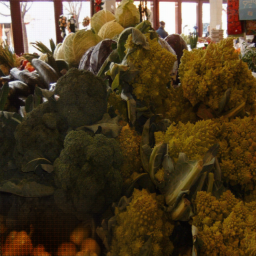}
     \includegraphics[width=0.48\textwidth]{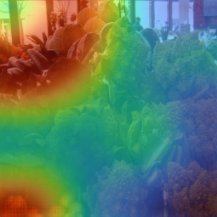}
     \caption*{\textbf{$f$'s answer}: red}
     \end{subfigure}
     \label{fig:exp:sem:a}
    \end{subfigure}
    
    \begin{subfigure}[b]{0.8\textwidth}
    \vspace{-0.5cm}
    \caption{\textbf{Question}: What color is the large central flower?}
     \begin{subfigure}[b]{0.5\textwidth}
     \includegraphics[width=0.48\textwidth]{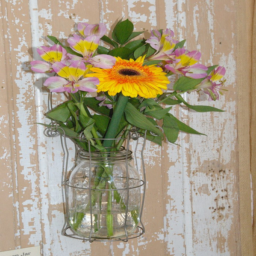}
     \includegraphics[width=0.48\textwidth]{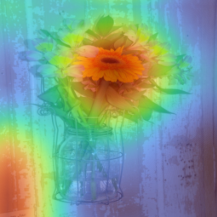}
     \caption*{\textbf{$f$'s answer}: yellow}
     \end{subfigure}
     \begin{subfigure}[b]{0.5\textwidth}
     \includegraphics[width=0.48\textwidth]{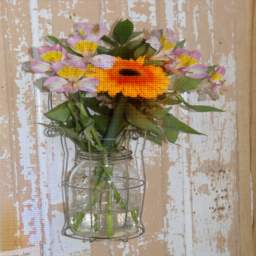}
     \includegraphics[width=0.48\textwidth]{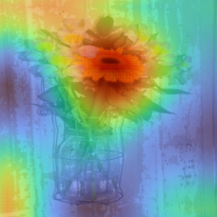}
     \caption*{\textbf{$f$'s answer}: orange}
     \end{subfigure}
     \label{fig:exp:sem:b}
    \end{subfigure}

    \begin{subfigure}[b]{0.8\textwidth}
    \vspace{-0.5cm}
    \caption{\textbf{Question}: What color is the bird?}
     \begin{subfigure}[b]{0.5\textwidth}
     \includegraphics[width=0.48\textwidth]{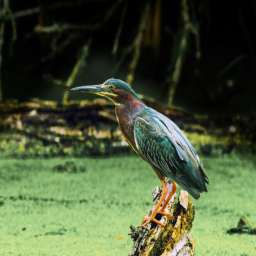}
     \includegraphics[width=0.48\textwidth]{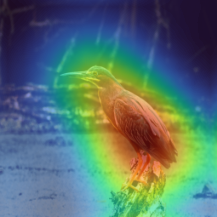}
     \caption*{\textbf{$f$'s answer}: blue and white}
     \end{subfigure}
     \begin{subfigure}[b]{0.5\textwidth}
     \includegraphics[width=0.48\textwidth]{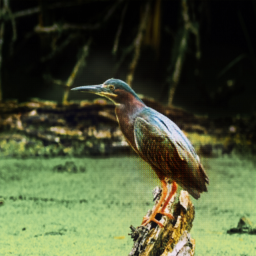}
     \includegraphics[width=0.48\textwidth]{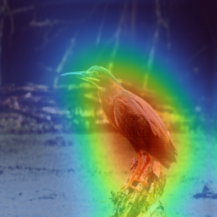}
     \caption*{\textbf{$f$'s answer}: brown and white}
     \end{subfigure}
     \label{fig:exp:sem:c}
    \end{subfigure}

    \begin{subfigure}[b]{0.8\textwidth}
    \vspace{-0.5cm}
    \caption{\textbf{Question}: What color is the dog?}
     \begin{subfigure}[b]{0.5\textwidth}
     \includegraphics[width=0.48\textwidth]{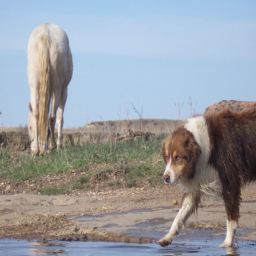}
     \includegraphics[width=0.48\textwidth]{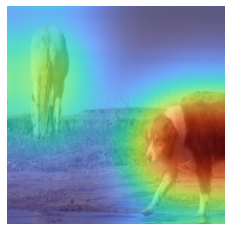}
     \caption*{\textbf{$f$'s answer}: brown}
     \end{subfigure}
     \begin{subfigure}[b]{0.5\textwidth}
     \includegraphics[width=0.48\textwidth]{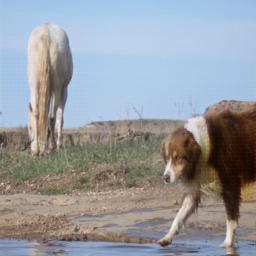}
     \includegraphics[width=0.48\textwidth]{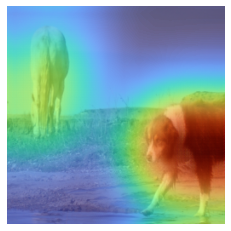}
     \caption*{\textbf{$f$'s answer}: brown}
     \end{subfigure}
     \label{fig:exp:sem:d}
    \end{subfigure}
    
    \begin{subfigure}[b]{0.8\textwidth}
    \vspace{-0.5cm}
    \caption{\textbf{Question}: What is the color scheme of the picture?}
     \begin{subfigure}[b]{0.5\textwidth}
     \includegraphics[width=0.48\textwidth]{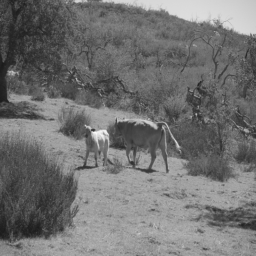}
     \includegraphics[width=0.48\textwidth]{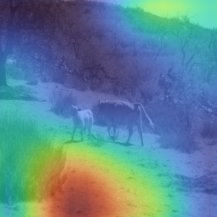}
     \caption*{\textbf{$f$'s answer}: black and white}
     \end{subfigure}
     \begin{subfigure}[b]{0.5\textwidth}
     \includegraphics[width=0.48\textwidth]{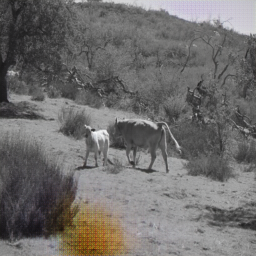}
     \includegraphics[width=0.48\textwidth]{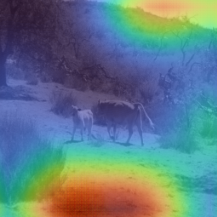}
     \caption*{\textbf{$f$'s answer}: orange and white}
     \end{subfigure}
     \label{fig:exp:sem:e}
    \end{subfigure}

    \caption{Example outputs of $\mathcal{G}$ for color-based questions from the VQAv1 \cite{antol2015vqa} validation set. Left: original image $I$ and the corresponding
attention map $M$. Right: Generated counterfactual image $I'$ and the corresponding attention map $M'$.}
    \label{fig:exp:sem}
\end{figure*}

Figure \ref{fig:exp:sem} illustrates qualitative results of $\mathcal{G}$ for color-based questions, where each row represents, from left to right, the original image $I$, the corresponding map $M$ and the generated counterfactual image $I'$ and its corresponding map $M'$. The rows \ref{fig:exp:sem:a} to \ref{fig:exp:sem:c} show examples of counterfactual images with different answers than their originals with realistic and understandable changes. As can be noticed in some examples such as Figures \ref{fig:exp:sem:a} and \ref{fig:exp:sem:b}, the VQA model $f$ slightly shifts its attention after the change is applied. This means, that theoretically, $f$ can give a different answer because of focusing on another region after the edit and not because of the edit itself.

For the rest of the samples, $f$ fails to alter their semantic meaning. Figure \ref{fig:exp:sem:d} depicts such an example, where $f$’s answer does not change when being provided with the counterfactual image. The cause of this failure is due to the difference between the original and counterfactual images is too small for $f$ to change the prediction. This because the clashing constraints that $\mathcal{G}$ has to obey. For example, $\mathcal{G}$ is required to change the output of $f$ but is at the same time restricted to apply as minimally as possible of changes. Another reason might be a wrong gaudiness of the attention map. Suppose the question-critical object accounts for a large portion of the image, or the question is about the image’s background. In that case, $\mathcal{G}$ often only edits those parts on which the attention mechanism focuses. As a result, the relevant image region is not modified in its entirety and thus $f$ cannot perceive a semantically meaningful change or its attention is shifted towards other regions of the question-critical object.

\begin{figure*}
\centering
    \begin{subfigure}[b]{0.8\textwidth}
    \vspace{-0.5cm}
    \caption{\textbf{Question}: What shape is the sign at the top of the post?}
     \begin{subfigure}[b]{0.5\textwidth}
     \includegraphics[width=0.48\textwidth]{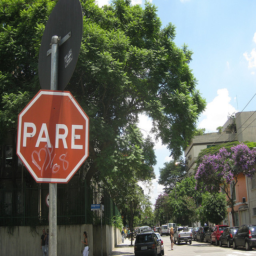}
     \includegraphics[width=0.48\textwidth]{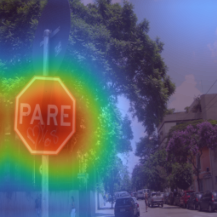}
     \caption*{\textbf{$f$'s answer}: triangle}
     \end{subfigure}
     \begin{subfigure}[b]{0.5\textwidth}
     \includegraphics[width=0.48\textwidth]{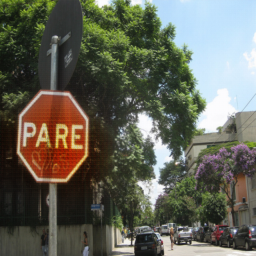}
     \includegraphics[width=0.48\textwidth]{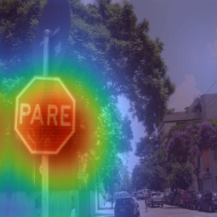}
     \caption*{\textbf{$f$'s answer}: square}
     \end{subfigure}
     \label{fig:exp:sem-s:a}
    \end{subfigure}
    
    \begin{subfigure}[b]{0.8\textwidth}
    \vspace{-0.5cm}
    \caption{\textbf{Question}: What shape is the kite at the top left of the image?}
     \begin{subfigure}[b]{0.5\textwidth}
     \includegraphics[width=0.48\textwidth]{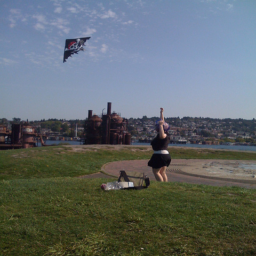}
     \includegraphics[width=0.48\textwidth]{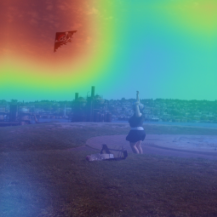}
     \caption*{\textbf{$f$'s answer}: triangle}
     \end{subfigure}
     \begin{subfigure}[b]{0.5\textwidth}
     \includegraphics[width=0.48\textwidth]{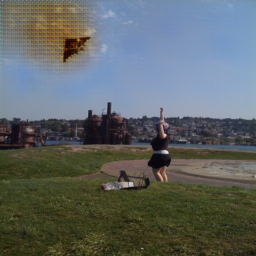}
     \includegraphics[width=0.48\textwidth]{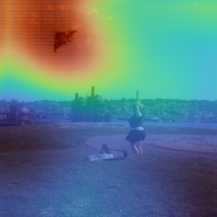}
     \caption*{\textbf{$f$'s answer}: round}
     \end{subfigure}
     \label{fig:exp:sem-s:b}
    \end{subfigure}

    \caption{Example outputs of $\mathcal{G}$ for shape-based questions from the VQAv1 [5] validation set. Left: original image $I$ and the corresponding
attention map $M$. Right: Generated counterfactual image $I'$ and the corresponding attention map $M'$.}
    \label{fig:exp:sem-s}
    \vspace{-0.3cm}
\end{figure*}

While $\mathcal{G}$ can generate semantically meaningful counterfactual images for a lot of color-based questions, it mostly fails in the context of shapes. Figure \ref{fig:exp:sem-s} shows two examples. While the VQA model predicts a different answer for the counterfactual than for the original image in both cases, the changes are not semantically meaningful from a human observer’s perspective. In example \ref{fig:exp:sem-s:a}, the object’s shape remains roughly unchanged, while the counterfactual generator slightly edits the sign’s color. Furthermore, $f$ predicts an incorrect answer regarding both the original and the counterfactual image. As $M$ indicates, the changes to the original image appear to be significant enough for $f$ to shift its focus slightly to the lower right portion of the sign when making an inference on the counterfactual image.

This shift seems to cause the model to change its prediction. Contrarily, the changes in example \ref{fig:exp:sem-s:b} are more dominant. $\mathcal{G}$ produces an artifact covering the kite and a segment of the sky surrounding it. As the $M$ show, $f$ focuses on the same area in both the original and the counterfactual image, but the artifact seems to cause it to change its answer. These two instances are exemplary for most of the counterfactual images for shape-based questions: the generator (i) applies only a few edits that are barely noticeable but they are more likely to change the answer of $f$ which is the task of $\mathcal{G}$ or (ii) produces artifacts that are not semantically meaningful for a human observer.

Both the examples in Figure \ref{fig:exp:sem} and Figure \ref{fig:exp:sem-s} show that $\mathcal{G}$’s edits vary depending on the questions and answers. Since the attention maps pose a strong constraint for $\mathcal{G}$, its edits heavily depend on them. If $M$ focuses on the question-critical object, such as in Figure \ref{fig:exp:sem:c}, $\mathcal{G}$ successfully modifies it. Contrarily, if $M$ focuses only on a small portion of the object in question (such as in Figure \ref{fig:exp:sem:e}), language conditioning does not have the desired effect. In these cases, $\mathcal{G}$ fails to modify the areas relevant to the question-answer pair sufficiently. Despite the failure to generate the counterfactual image, $\mathcal{G}$ provides an understandable interpretability to the behaviour and result of the VQA model for a particular pair (Image, Question).

\begin{figure*}
    \centering
     \begin{subfigure}[b]{0.75\textwidth}
     \caption{\textbf{Question}: What color is the ball? \textbf{MUTAN answer}: orange.}
     \includegraphics[width=0.3\textwidth]{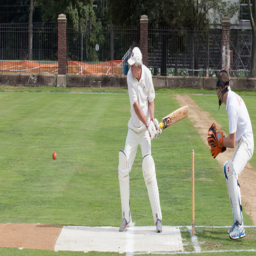}
     \hfill
     \includegraphics[width=0.3\textwidth]{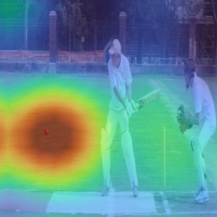}
     \hfill
     \includegraphics[width=0.3\textwidth]{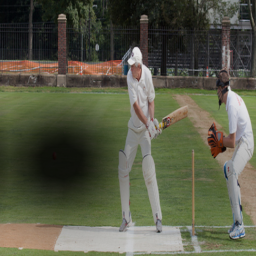}
     
     \label{fig:exp:att:a}
     \end{subfigure}
     
     \begin{subfigure}[b]{0.75\textwidth}
     \caption{\textbf{Question}: What color is the bus? \textbf{MUTAN answer}: red.}
     \includegraphics[width=0.3\textwidth]{Figures/b_1.png}
     \hfill
     \includegraphics[width=0.3\textwidth]{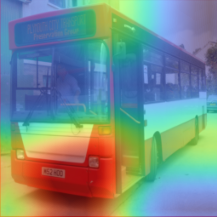}
     \hfill
     \includegraphics[width=0.3\textwidth]{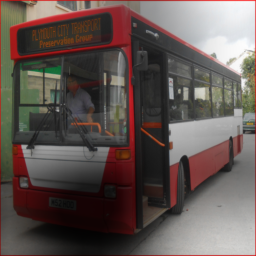}
     
     \label{fig:exp:att:b}
     \end{subfigure}

     \begin{subfigure}[b]{0.75\textwidth}
     \caption{\textbf{Question}: What color are the flags? \textbf{MUTAN answer}: Red and white.}
     \includegraphics[width=0.3\textwidth]{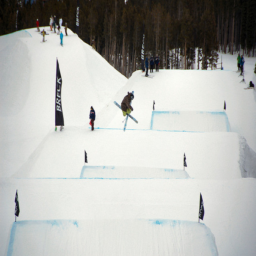}
     \hfill
     \includegraphics[width=0.3\textwidth]{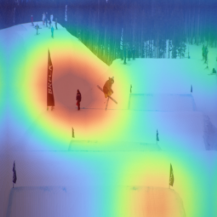}
     \hfill
     \includegraphics[width=0.3\textwidth]{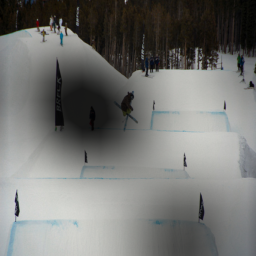}
     
     \label{fig:exp:att:c}
     \end{subfigure}
    \caption{Example outputs of the Grad-CAM algorithm applied to MUTAN for color-based questions. Left: original image. Center: Interpolated attention map projected on the original image. Right: The background image.}
    \label{fig:exp:att}
    \vspace{-0.2cm}
\end{figure*}

\paragraph{Question-critical object (RQ2):}
One of the main aims of \appname~ is to edit only the question-critical object. This is controlled by the attention map $M$, where the generator $\mathcal{G}$ is restricted to apply changes predominantly in the image regions on which $M$ focuses. Figure~\ref{fig:exp:att} depicts, for three example samples, the original image $I$ (Left), the question $Q$, the answer $A$ given by MUTAN ($f$) and the interpolation of the map $M$ with $I$ (Center). The right image is the \emph{background} obtained by computing $(\mathbf{1}-M) \odot I$, where $\mathbf{1}$ denotes all-ones matrix and $\odot$ is the element-wise multiplication. 

In Figure \ref{fig:exp:att:a}, the question is about the ball and as shown in the interpolated map, $f$ focuses specifically on the ball's region, which makes $\mathcal{G}$ restricted to make changes only on that region. When the region of interest is large and/or sparse such as in Figure \ref{fig:exp:att:b}, $f$ might not focus on the entire question-critical region but only a portion of it, which is sufficient to answer the question. Figure \ref{fig:exp:att:c} indicates that $f$ wrongly answered the question as \emph{red and white} but the correct answer is clearly \emph{black}. The obtained map $M$ explains that $f$ was focusing on a different region. Consequently, $\mathcal{G}$ applies the changes on the wrong region. The user can understand the behaviour of the model based on this generated image, which is an edit to the original one w.r.t a wrong region.

According to our user study,  $\mathcal{G}$ applied the changes on the critical object in $50.1\%$ of the samples. For $32.2\%$ of the samples, the changes were applied on (1) completely different region, (2) the region of interest and other irrelevant regions or (3) a small part of the region of interest. For the remaining $17.7\%$, the users couldn't determine whether the changes were applied on the question-critical object or not. From these results, we can derive three main patterns of the attention mechanism:
\begin{enumerate}[leftmargin=*]
    \item If the object relevant to answering a question is relatively small compared to the rest of the image, the attention mechanism focuses on it completely in most cases. In other words, the computed intensities are higher for pixels belonging to the object than for the rest of the pixels. Under these circumstances, the generator can make larger changes to the entire object than to the rest of the image. 
    
    \item Contrarily, if the object is very large or MUTAN pays attention to the background, the projection usually focuses only on a part of it. Consequently, the information that the generator receives allows it to apply more significant changes to a segment of the object or the background than to the rest of it. 
    
    \item If MUTAN makes an incorrect prediction, this is often reflected by the projection not focusing on the question-critical object, but another element of the image, such as in Figure \ref{fig:exp:att:c}.
    
\end{enumerate}

In all these patterns, the applied changes of $I$ w.r.t $M$ is supposed to change the answer of $f$ for the same question $Q$. This is because $M$ is indicating where $f$ is focusing to answer $Q$ given $I$. However, when the VQA focuses on an irrelevant region in $I$, the applied changes might change the visual semantic of this irrelevant region such that when feeding $f$ with $I'$ and $Q$, $f$ focuses on a different region than it did in $I$. This different region can also be the correct region.

\begin{figure}[H]
    \centering
    \includegraphics[width = \linewidth]{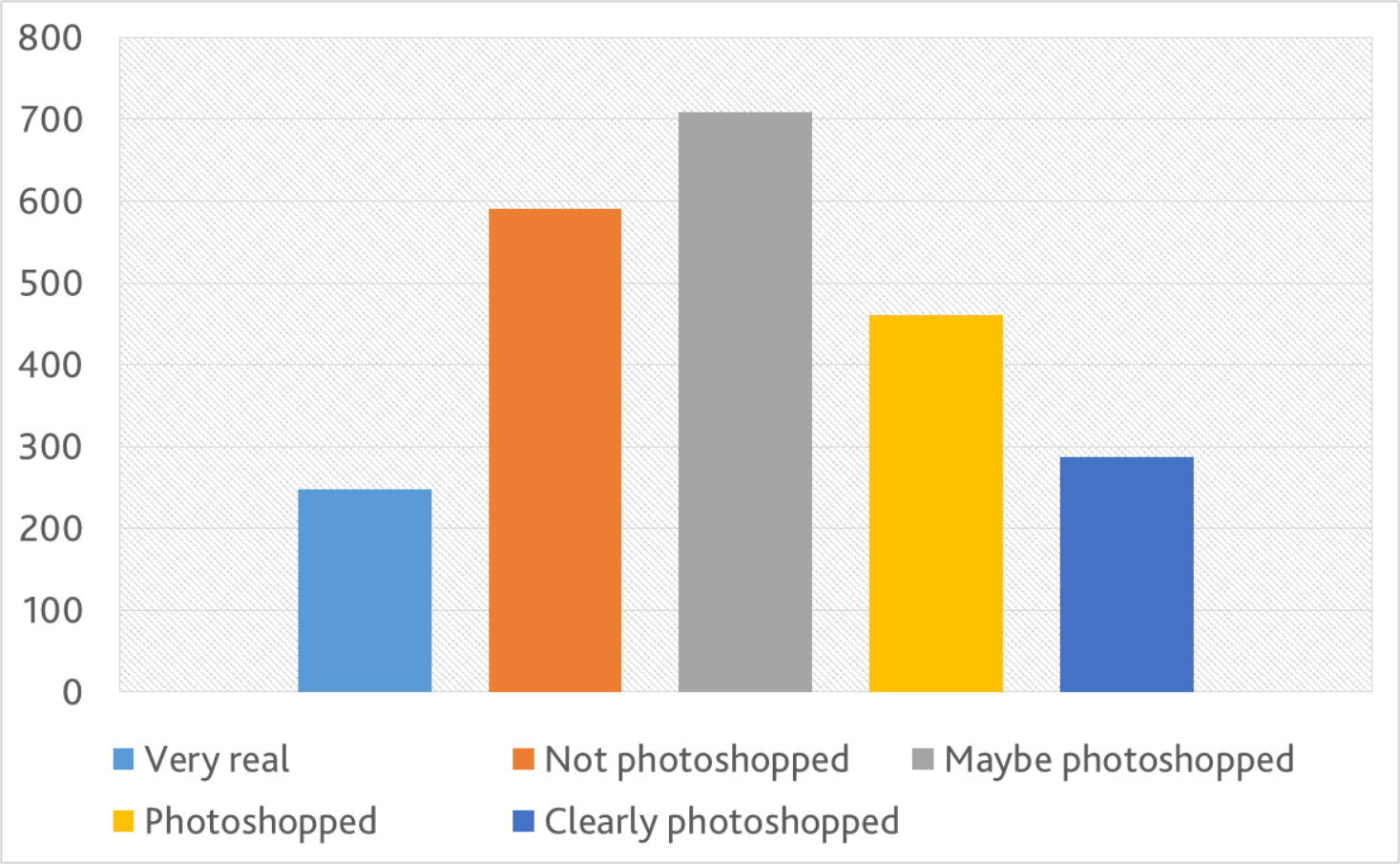}
    \caption{Frequency histogram of participants' answers to the question: ``\emph{Does the picture look photoshopped?}''}
    \label{fig:exp:real}
    \vspace{-0.3cm}
\end{figure}

\paragraph{Realism (RQ3):}
Generating realistic counterfactual images is very important to interpret the result of VQA models to users. As shown in Figure \ref{fig:exp:sem} and Figure \ref{fig:exp:sem-s}, the degree of realism varies depending on the necessary edit that changes $f$' answer and on the size difference of the question-critical region to the focused object. This is reflected also in the result of our user study that is demonstrated in Figure \ref{fig:exp:real}, where the users were presented only with the generated images and asked ``\emph{Does the picture look photoshopped (any noticeable edit or distortion)?}''. As Figure \ref{fig:exp:real} indicates, the answers of the users vary depending on the generated images. When presenting the corresponding original image together with the generated one and asking ``\emph{Which of the images is the original?}'', the participants could correctly distinguish between the original and the generated one in $ \thicksim 67.2\%$ of the presented samples. In $ \thicksim 12.9\%$ of the images, the participants selected the generated image as the original one and in the rest of image ($\thicksim 19.9\%$), the participants couldn't decide which of the images is the original and which one is the generated. This result indicates that $\mathcal{G}$ could trick the human participants by generating counterfactual which look extremely realistic.

\paragraph{Minimality of Image edits:}

To evaluate $\mathcal{G}$'s performance on generating counterfactual images with minimum edits, we computed $\ell_1$-norm across both the training and the validation set. This measures the magnitude of changes in the counterfactual image relative to the original one, where lower values indicate fewer changes.

Table \ref{tab:exp:min} summarizes the results of this evaluation, where the mean (denoted $\mu$) and standard deviation (denoted $\sigma$) values are calculated for different splits of both the training and the validation sets. The first three columns represent the values computed across the entire dataset and for color-based and shape-based questions. The remaining six columns contain the same computations for the portion of pairs of original and counterfactual images for which $\mathcal{G}$ predicts distinct or equal answers, respectively.

\begin{table*}[ht]
    \centering
    \begin{tabular}{cc|ccc|ccc|ccc}
    \myhline
    \multicolumn{2}{c|}{\multirow{2}{*}{}} & \multirow{2}{*}{\textbf{All}} & \multirow{2}{*}{\textbf{Color}} & \multirow{2}{*}{\textbf{Shape}} & \multicolumn{3}{c|}{\textbf{Same VQA Answers}} & \multicolumn{3}{c}{\textbf{Different VQA Answers}}\\
    \cline{6-11}
    \multicolumn{2}{c|}{} &  &  & & \textbf{All} & \textbf{Color} & \textbf{Shape} & \textbf{All} & \textbf{Color} & \textbf{Shape}  \\
    \hline
    \multirow{2}{*}{Training set} & $\mu$ & $0.0175$ & $\mathbf{0.0174}$ & $0.0207$ & $0.0177$ & $\mathbf{0.0176}$ & $0.0212$ & $0.0173$ & $\mathbf{0.0172}$ & $0.0195$\\
     & $\sigma$ & $\mathbf{0.0039}$ & $\mathbf{0.0039}$ & $0.0048$ & $0.0040$ & $\mathbf{0.0039}$ & $0.0049$ & $0.0039$ & $\mathbf{0.0038}$ & $0.0046$\\
     \hline
     \multirow{2}{*}{Validation set} & $\mu$ & $0.0175$ & $\mathbf{0.0174}$ & $0.0208$ & $0.0177$ & $\mathbf{0.0176}$ & $0.0212$ & $\mathbf{0.0173}$ & $\mathbf{0.0173}$ & $0.0198$\\
      & $\sigma$ & $0.0041$ & $\mathbf{0.004}$ & $0.0047$ & $0.0041$  & $0.0041$ & $0.0048$ & $\mathbf{0.0038}$ & $\mathbf{0.0038}$ & $0.0042$\\
    \myhline
    \end{tabular}
    \caption{Mean ($\mu$) and standard deviation ($\sigma$) of the $\ell_1$-norm computed across the training and validation set and split across categories}
    \label{tab:exp:min}
    \vspace{-0.4cm}
\end{table*}

The results of Table \ref{tab:exp:min} indicate that $\mathcal{G}$ applied fewer changes when it comes to color-based questions compared to shape-related questions. This observation applies to both the training and the validation set and across all splits. Moreover, overall, $\mathcal{G}$ applied fewer changes in those cases where $f$’s predictions regarding the original and counterfactual image were distinct than if they were equal. This indicates that $\mathcal{G}$ changed the original image to the maximum possible level but without successfully changing $f$'s answer.

\section{Conclusion}
\label{sec:conc}

In this paper, we introduced \appname~, a GAN-based approach to interpret the output of VQA models by generating counterfactual images to drive the VQA model outputting different a different answer (\emph{\textbf{RQ1}}). \appname~ is a modified implementation of LingUNet with incorporating a Grad-CAM-based attention mechanism that determines each pixel’s importance regarding the VQA model’s decision making process. With this, the counterfactual generator learns to apply modifications in an image predominantly to question-critical objects, while retaining the rest of the image (\emph{\textbf{RQ2}}). The obtained results indicate that using an attention mechanism is an appropriate means to guide the modification process. Furthermore, the quality of the counterfactual images depended to a large extent on the attention maps. Extensive experiments on the challenging VQAv1 dataset have demonstrated that \appname~ achieves encouraging results for color questions by generating realistic counterfactual images (\emph{\textbf{RQ3}}).

For future work, we will train \appname~ on a larger, more diverse dataset such as VQAv2 dataset, which contains multiple images per question rather than only a single one as in the VQAv1 dataset. Moreover, using an attention mechanism that focuses on the question-critical objects more accurately could also significantly improve the interpretability capabilities of \appname. To this end, we plan to employ super-pixel segmentation to extract concepts (e.g. color, texture, or a group of similar segments)  and uses the Shapley Value algorithm to determine each concept’s contribution to a DNN’s decision. The generator will then be trained to alter  the most important concept(s) in an image rather than providing it an attention map. Moreover, replacing an entire instance of a concept rather than editing an image on a pixel-by-pixel level could pave the way for semantic changes even larger than altering shapes.

\bibliographystyle{ACM-Reference-Format}
\bibliography{bibliography}


\begin{thebibliography}{53}


\ifx \showCODEN    \undefined \def \showCODEN     #1{\unskip}     \fi
\ifx \showDOI      \undefined \def \showDOI       #1{#1}\fi
\ifx \showISBNx    \undefined \def \showISBNx     #1{\unskip}     \fi
\ifx \showISBNxiii \undefined \def \showISBNxiii  #1{\unskip}     \fi
\ifx \showISSN     \undefined \def \showISSN      #1{\unskip}     \fi
\ifx \showLCCN     \undefined \def \showLCCN      #1{\unskip}     \fi
\ifx \shownote     \undefined \def \shownote      #1{#1}          \fi
\ifx \showarticletitle \undefined \def \showarticletitle #1{#1}   \fi
\ifx \showURL      \undefined \def \showURL       {\relax}        \fi
\providecommand\bibfield[2]{#2}
\providecommand\bibinfo[2]{#2}
\providecommand\natexlab[1]{#1}
\providecommand\showeprint[2][]{arXiv:#2}

\bibitem[\protect\citeauthoryear{Anderson, He, Buehler, Teney, Johnson, Gould,
  and Zhang}{Anderson et~al\mbox{.}}{2018}]%
        {anderson2018bottom}
\bibfield{author}{\bibinfo{person}{Peter Anderson}, \bibinfo{person}{Xiaodong
  He}, \bibinfo{person}{Chris Buehler}, \bibinfo{person}{Damien Teney},
  \bibinfo{person}{Mark Johnson}, \bibinfo{person}{Stephen Gould}, {and}
  \bibinfo{person}{Lei Zhang}.} \bibinfo{year}{2018}\natexlab{}.
\newblock \showarticletitle{Bottom-up and top-down attention for image
  captioning and visual question answering}. In
  \bibinfo{booktitle}{\emph{Proceedings of the IEEE conference on computer
  vision and pattern recognition}}. \bibinfo{pages}{6077--6086}.
\newblock


\bibitem[\protect\citeauthoryear{Antol, Agrawal, Lu, Mitchell, Batra, Zitnick,
  and Parikh}{Antol et~al\mbox{.}}{2015}]%
        {antol2015vqa}
\bibfield{author}{\bibinfo{person}{Stanislaw Antol}, \bibinfo{person}{Aishwarya
  Agrawal}, \bibinfo{person}{Jiasen Lu}, \bibinfo{person}{Margaret Mitchell},
  \bibinfo{person}{Dhruv Batra}, \bibinfo{person}{C~Lawrence Zitnick}, {and}
  \bibinfo{person}{Devi Parikh}.} \bibinfo{year}{2015}\natexlab{}.
\newblock \showarticletitle{Vqa: Visual question answering}. In
  \bibinfo{booktitle}{\emph{Proceedings of the IEEE international conference on
  computer vision}}. \bibinfo{pages}{2425--2433}.
\newblock


\bibitem[\protect\citeauthoryear{Ben-Younes, Cadene, Cord, and
  Thome}{Ben-Younes et~al\mbox{.}}{2017}]%
        {ben2017mutan}
\bibfield{author}{\bibinfo{person}{Hedi Ben-Younes}, \bibinfo{person}{R{\'e}mi
  Cadene}, \bibinfo{person}{Matthieu Cord}, {and} \bibinfo{person}{Nicolas
  Thome}.} \bibinfo{year}{2017}\natexlab{}.
\newblock \showarticletitle{Mutan: Multimodal tucker fusion for visual question
  answering}. In \bibinfo{booktitle}{\emph{Proceedings of the IEEE
  international conference on computer vision}}. \bibinfo{pages}{2612--2620}.
\newblock


\bibitem[\protect\citeauthoryear{Bergholm}{Bergholm}{1987}]%
        {bergholm1987edge}
\bibfield{author}{\bibinfo{person}{Fredrik Bergholm}.}
  \bibinfo{year}{1987}\natexlab{}.
\newblock \showarticletitle{Edge focusing}.
\newblock \bibinfo{journal}{\emph{IEEE transactions on pattern analysis and
  machine intelligence}} \bibinfo{number}{6} (\bibinfo{year}{1987}),
  \bibinfo{pages}{726--741}.
\newblock


\bibitem[\protect\citeauthoryear{Chakraborty, Tomsett, Raghavendra, Harborne,
  Alzantot, Cerutti, Srivastava, Preece, Julier, Rao,
  et~al\mbox{.}}{Chakraborty et~al\mbox{.}}{2017}]%
        {chakraborty2017interpretability}
\bibfield{author}{\bibinfo{person}{Supriyo Chakraborty},
  \bibinfo{person}{Richard Tomsett}, \bibinfo{person}{Ramya Raghavendra},
  \bibinfo{person}{Daniel Harborne}, \bibinfo{person}{Moustafa Alzantot},
  \bibinfo{person}{Federico Cerutti}, \bibinfo{person}{Mani Srivastava},
  \bibinfo{person}{Alun Preece}, \bibinfo{person}{Simon Julier},
  \bibinfo{person}{Raghuveer~M Rao}, {et~al\mbox{.}}}
  \bibinfo{year}{2017}\natexlab{}.
\newblock \showarticletitle{Interpretability of deep learning models: A survey
  of results}. In \bibinfo{booktitle}{\emph{2017 IEEE smartworld, ubiquitous
  intelligence \& computing, advanced \& trusted computed, scalable computing
  \& communications, cloud \& big data computing, Internet of people and smart
  city innovation (smartworld/SCALCOM/UIC/ATC/CBDcom/IOP/SCI)}}. IEEE,
  \bibinfo{pages}{1--6}.
\newblock


\bibitem[\protect\citeauthoryear{Chen, Yan, Xiao, Zhang, Pu, and Zhuang}{Chen
  et~al\mbox{.}}{2020}]%
        {chen2020counterfactual}
\bibfield{author}{\bibinfo{person}{Long Chen}, \bibinfo{person}{Xin Yan},
  \bibinfo{person}{Jun Xiao}, \bibinfo{person}{Hanwang Zhang},
  \bibinfo{person}{Shiliang Pu}, {and} \bibinfo{person}{Yueting Zhuang}.}
  \bibinfo{year}{2020}\natexlab{}.
\newblock \showarticletitle{Counterfactual samples synthesizing for robust
  visual question answering}. In \bibinfo{booktitle}{\emph{Proceedings of the
  IEEE/CVF Conference on Computer Vision and Pattern Recognition}}.
  \bibinfo{pages}{10800--10809}.
\newblock


\bibitem[\protect\citeauthoryear{Das, Agrawal, Zitnick, Parikh, and Batra}{Das
  et~al\mbox{.}}{2017}]%
        {das2017human}
\bibfield{author}{\bibinfo{person}{Abhishek Das}, \bibinfo{person}{Harsh
  Agrawal}, \bibinfo{person}{Larry Zitnick}, \bibinfo{person}{Devi Parikh},
  {and} \bibinfo{person}{Dhruv Batra}.} \bibinfo{year}{2017}\natexlab{}.
\newblock \showarticletitle{Human attention in visual question answering: Do
  humans and deep networks look at the same regions?}
\newblock \bibinfo{journal}{\emph{Computer Vision and Image Understanding}}
  \bibinfo{volume}{163} (\bibinfo{year}{2017}), \bibinfo{pages}{90--100}.
\newblock


\bibitem[\protect\citeauthoryear{Denton, Hutchinson, Mitchell, Gebru, and
  Zaldivar}{Denton et~al\mbox{.}}{2019}]%
        {denton2019image}
\bibfield{author}{\bibinfo{person}{Emily Denton}, \bibinfo{person}{Ben
  Hutchinson}, \bibinfo{person}{Margaret Mitchell}, \bibinfo{person}{Timnit
  Gebru}, {and} \bibinfo{person}{Andrew Zaldivar}.}
  \bibinfo{year}{2019}\natexlab{}.
\newblock \showarticletitle{Image counterfactual sensitivity analysis for
  detecting unintended bias}.
\newblock \bibinfo{journal}{\emph{arXiv preprint arXiv:1906.06439}}
  (\bibinfo{year}{2019}).
\newblock


\bibitem[\protect\citeauthoryear{Doshi-Velez and Kim}{Doshi-Velez and
  Kim}{2017}]%
        {doshi2017towards}
\bibfield{author}{\bibinfo{person}{Finale Doshi-Velez} {and}
  \bibinfo{person}{Been Kim}.} \bibinfo{year}{2017}\natexlab{}.
\newblock \showarticletitle{Towards a rigorous science of interpretable machine
  learning}.
\newblock \bibinfo{journal}{\emph{arXiv preprint arXiv:1702.08608}}
  (\bibinfo{year}{2017}).
\newblock


\bibitem[\protect\citeauthoryear{Fern{\'a}ndez-Lor{\'\i}a, Provost, and
  Han}{Fern{\'a}ndez-Lor{\'\i}a et~al\mbox{.}}{2020}]%
        {fernandez2020explaining}
\bibfield{author}{\bibinfo{person}{Carlos Fern{\'a}ndez-Lor{\'\i}a},
  \bibinfo{person}{Foster Provost}, {and} \bibinfo{person}{Xintian Han}.}
  \bibinfo{year}{2020}\natexlab{}.
\newblock \showarticletitle{Explaining data-driven decisions made by AI
  systems: the counterfactual approach}.
\newblock \bibinfo{journal}{\emph{arXiv preprint arXiv:2001.07417}}
  (\bibinfo{year}{2020}).
\newblock


\bibitem[\protect\citeauthoryear{Geman, Geman, Hallonquist, and Younes}{Geman
  et~al\mbox{.}}{2015}]%
        {geman2015visual}
\bibfield{author}{\bibinfo{person}{Donald Geman}, \bibinfo{person}{Stuart
  Geman}, \bibinfo{person}{Neil Hallonquist}, {and} \bibinfo{person}{Laurent
  Younes}.} \bibinfo{year}{2015}\natexlab{}.
\newblock \showarticletitle{Visual turing test for computer vision systems}.
\newblock \bibinfo{journal}{\emph{Proceedings of the National Academy of
  Sciences}} \bibinfo{volume}{112}, \bibinfo{number}{12}
  (\bibinfo{year}{2015}), \bibinfo{pages}{3618--3623}.
\newblock


\bibitem[\protect\citeauthoryear{Gilpin, Bau, Yuan, Bajwa, Specter, and
  Kagal}{Gilpin et~al\mbox{.}}{2018}]%
        {gilpin2018explaining}
\bibfield{author}{\bibinfo{person}{Leilani~H Gilpin}, \bibinfo{person}{David
  Bau}, \bibinfo{person}{Ben~Z Yuan}, \bibinfo{person}{Ayesha Bajwa},
  \bibinfo{person}{Michael Specter}, {and} \bibinfo{person}{Lalana Kagal}.}
  \bibinfo{year}{2018}\natexlab{}.
\newblock \showarticletitle{Explaining explanations: An overview of
  interpretability of machine learning}. In \bibinfo{booktitle}{\emph{2018 IEEE
  5th International Conference on data science and advanced analytics (DSAA)}}.
  IEEE, \bibinfo{pages}{80--89}.
\newblock


\bibitem[\protect\citeauthoryear{Gomez, Holter, Yuan, and Bertini}{Gomez
  et~al\mbox{.}}{2020}]%
        {gomez2020vice}
\bibfield{author}{\bibinfo{person}{Oscar Gomez}, \bibinfo{person}{Steffen
  Holter}, \bibinfo{person}{Jun Yuan}, {and} \bibinfo{person}{Enrico Bertini}.}
  \bibinfo{year}{2020}\natexlab{}.
\newblock \showarticletitle{ViCE: visual counterfactual explanations for
  machine learning models}. In \bibinfo{booktitle}{\emph{Proceedings of the
  25th International Conference on Intelligent User Interfaces}}.
  \bibinfo{pages}{531--535}.
\newblock


\bibitem[\protect\citeauthoryear{Goodfellow, Pouget-Abadie, Mirza, Xu,
  Warde-Farley, Ozair, Courville, and Bengio}{Goodfellow et~al\mbox{.}}{2020}]%
        {goodfellow2020generative}
\bibfield{author}{\bibinfo{person}{Ian Goodfellow}, \bibinfo{person}{Jean
  Pouget-Abadie}, \bibinfo{person}{Mehdi Mirza}, \bibinfo{person}{Bing Xu},
  \bibinfo{person}{David Warde-Farley}, \bibinfo{person}{Sherjil Ozair},
  \bibinfo{person}{Aaron Courville}, {and} \bibinfo{person}{Yoshua Bengio}.}
  \bibinfo{year}{2020}\natexlab{}.
\newblock \showarticletitle{Generative adversarial networks}.
\newblock \bibinfo{journal}{\emph{Commun. ACM}} \bibinfo{volume}{63},
  \bibinfo{number}{11} (\bibinfo{year}{2020}), \bibinfo{pages}{139--144}.
\newblock


\bibitem[\protect\citeauthoryear{Goyal, Khot, Summers-Stay, Batra, and
  Parikh}{Goyal et~al\mbox{.}}{2017}]%
        {goyal2017making}
\bibfield{author}{\bibinfo{person}{Yash Goyal}, \bibinfo{person}{Tejas Khot},
  \bibinfo{person}{Douglas Summers-Stay}, \bibinfo{person}{Dhruv Batra}, {and}
  \bibinfo{person}{Devi Parikh}.} \bibinfo{year}{2017}\natexlab{}.
\newblock \showarticletitle{Making the v in vqa matter: Elevating the role of
  image understanding in visual question answering}. In
  \bibinfo{booktitle}{\emph{Proceedings of the IEEE Conference on Computer
  Vision and Pattern Recognition}}. \bibinfo{pages}{6904--6913}.
\newblock


\bibitem[\protect\citeauthoryear{Goyal, Wu, Ernst, Batra, Parikh, and
  Lee}{Goyal et~al\mbox{.}}{2019}]%
        {goyal2019counterfactual}
\bibfield{author}{\bibinfo{person}{Yash Goyal}, \bibinfo{person}{Ziyan Wu},
  \bibinfo{person}{Jan Ernst}, \bibinfo{person}{Dhruv Batra},
  \bibinfo{person}{Devi Parikh}, {and} \bibinfo{person}{Stefan Lee}.}
  \bibinfo{year}{2019}\natexlab{}.
\newblock \showarticletitle{Counterfactual visual explanations}. In
  \bibinfo{booktitle}{\emph{International Conference on Machine Learning}}.
  PMLR, \bibinfo{pages}{2376--2384}.
\newblock


\bibitem[\protect\citeauthoryear{Gupta}{Gupta}{2017}]%
        {gupta2017survey}
\bibfield{author}{\bibinfo{person}{Akshay~Kumar Gupta}.}
  \bibinfo{year}{2017}\natexlab{}.
\newblock \showarticletitle{Survey of visual question answering: Datasets and
  techniques}.
\newblock \bibinfo{journal}{\emph{arXiv preprint arXiv:1705.03865}}
  (\bibinfo{year}{2017}).
\newblock


\bibitem[\protect\citeauthoryear{Hendricks, Hu, Darrell, and Akata}{Hendricks
  et~al\mbox{.}}{2018}]%
        {hendricks2018generating}
\bibfield{author}{\bibinfo{person}{Lisa~Anne Hendricks},
  \bibinfo{person}{Ronghang Hu}, \bibinfo{person}{Trevor Darrell}, {and}
  \bibinfo{person}{Zeynep Akata}.} \bibinfo{year}{2018}\natexlab{}.
\newblock \showarticletitle{Generating counterfactual explanations with natural
  language}.
\newblock \bibinfo{journal}{\emph{arXiv preprint arXiv:1806.09809}}
  (\bibinfo{year}{2018}).
\newblock


\bibitem[\protect\citeauthoryear{Isola, Zhu, Zhou, and Efros}{Isola
  et~al\mbox{.}}{2017}]%
        {isola2017image}
\bibfield{author}{\bibinfo{person}{Phillip Isola}, \bibinfo{person}{Jun-Yan
  Zhu}, \bibinfo{person}{Tinghui Zhou}, {and} \bibinfo{person}{Alexei~A
  Efros}.} \bibinfo{year}{2017}\natexlab{}.
\newblock \showarticletitle{Image-to-image translation with conditional
  adversarial networks}. In \bibinfo{booktitle}{\emph{Proceedings of the IEEE
  conference on computer vision and pattern recognition}}.
  \bibinfo{pages}{1125--1134}.
\newblock


\bibitem[\protect\citeauthoryear{Kim, Khanna, and Koyejo}{Kim
  et~al\mbox{.}}{2016}]%
        {kim2016examples}
\bibfield{author}{\bibinfo{person}{Been Kim}, \bibinfo{person}{Rajiv Khanna},
  {and} \bibinfo{person}{Oluwasanmi~O Koyejo}.}
  \bibinfo{year}{2016}\natexlab{}.
\newblock \showarticletitle{Examples are not enough, learn to criticize!
  criticism for interpretability}.
\newblock \bibinfo{journal}{\emph{Advances in neural information processing
  systems}}  \bibinfo{volume}{29} (\bibinfo{year}{2016}).
\newblock


\bibitem[\protect\citeauthoryear{Krishna, Zhu, Groth, Johnson, Hata, Kravitz,
  Chen, Kalantidis, Li, Shamma, et~al\mbox{.}}{Krishna et~al\mbox{.}}{2016}]%
        {krishna2016visual}
\bibfield{author}{\bibinfo{person}{Ranjay Krishna}, \bibinfo{person}{Yuke Zhu},
  \bibinfo{person}{Oliver Groth}, \bibinfo{person}{Justin Johnson},
  \bibinfo{person}{Kenji Hata}, \bibinfo{person}{Joshua Kravitz},
  \bibinfo{person}{Stephanie Chen}, \bibinfo{person}{Yannis Kalantidis},
  \bibinfo{person}{Li-Jia Li}, \bibinfo{person}{David~A Shamma},
  {et~al\mbox{.}}} \bibinfo{year}{2016}\natexlab{}.
\newblock \showarticletitle{Visual genome: Connecting language and vision using
  crowdsourced dense image annotations}.
\newblock \bibinfo{journal}{\emph{arXiv preprint arXiv:1602.07332}}
  (\bibinfo{year}{2016}).
\newblock


\bibitem[\protect\citeauthoryear{Li, Fu, Yu, Mei, and Luo}{Li
  et~al\mbox{.}}{2018}]%
        {li2018tell}
\bibfield{author}{\bibinfo{person}{Qing Li}, \bibinfo{person}{Jianlong Fu},
  \bibinfo{person}{Dongfei Yu}, \bibinfo{person}{Tao Mei}, {and}
  \bibinfo{person}{Jiebo Luo}.} \bibinfo{year}{2018}\natexlab{}.
\newblock \showarticletitle{Tell-and-Answer: Towards Explainable Visual
  Question Answering using Attributes and Captions}. In
  \bibinfo{booktitle}{\emph{Proceedings of the 2018 Conference on Empirical
  Methods in Natural Language Processing}}. \bibinfo{pages}{1338--1346}.
\newblock


\bibitem[\protect\citeauthoryear{Lin, Sekar, and Fanti}{Lin
  et~al\mbox{.}}{2020}]%
        {lin2020spectral}
\bibfield{author}{\bibinfo{person}{Zinan Lin}, \bibinfo{person}{Vyas Sekar},
  {and} \bibinfo{person}{Giulia Fanti}.} \bibinfo{year}{2020}\natexlab{}.
\newblock \showarticletitle{Why Spectral Normalization Stabilizes GANs:
  Analysis and Improvements}.
\newblock \bibinfo{journal}{\emph{arXiv preprint arXiv:2009.02773}}
  (\bibinfo{year}{2020}).
\newblock


\bibitem[\protect\citeauthoryear{Malinowski and Fritz}{Malinowski and
  Fritz}{2014}]%
        {malinowski2014multi}
\bibfield{author}{\bibinfo{person}{Mateusz Malinowski} {and}
  \bibinfo{person}{Mario Fritz}.} \bibinfo{year}{2014}\natexlab{}.
\newblock \showarticletitle{A multi-world approach to question answering about
  real-world scenes based on uncertain input}.
\newblock \bibinfo{journal}{\emph{Advances in neural information processing
  systems}}  \bibinfo{volume}{27} (\bibinfo{year}{2014}),
  \bibinfo{pages}{1682--1690}.
\newblock


\bibitem[\protect\citeauthoryear{Misra, Bennett, Blukis, Niklasson, Shatkhin,
  and Artzi}{Misra et~al\mbox{.}}{2018}]%
        {misra2018mapping}
\bibfield{author}{\bibinfo{person}{Dipendra Misra}, \bibinfo{person}{Andrew
  Bennett}, \bibinfo{person}{Valts Blukis}, \bibinfo{person}{Eyvind Niklasson},
  \bibinfo{person}{Max Shatkhin}, {and} \bibinfo{person}{Yoav Artzi}.}
  \bibinfo{year}{2018}\natexlab{}.
\newblock \showarticletitle{Mapping Instructions to Actions in 3D Environments
  with Visual Goal Prediction}. In \bibinfo{booktitle}{\emph{Proceedings of the
  2018 Conference on Empirical Methods in Natural Language Processing}}.
  \bibinfo{pages}{2667--2678}.
\newblock


\bibitem[\protect\citeauthoryear{Miyato, Kataoka, Koyama, and Yoshida}{Miyato
  et~al\mbox{.}}{2018}]%
        {miyato2018spectral}
\bibfield{author}{\bibinfo{person}{Takeru Miyato}, \bibinfo{person}{Toshiki
  Kataoka}, \bibinfo{person}{Masanori Koyama}, {and} \bibinfo{person}{Yuichi
  Yoshida}.} \bibinfo{year}{2018}\natexlab{}.
\newblock \showarticletitle{Spectral Normalization for Generative Adversarial
  Networks}. In \bibinfo{booktitle}{\emph{International Conference on Learning
  Representations}}.
\newblock


\bibitem[\protect\citeauthoryear{Montavon, Lapuschkin, Binder, Samek, and
  M{\"u}ller}{Montavon et~al\mbox{.}}{2017}]%
        {montavon2017explaining}
\bibfield{author}{\bibinfo{person}{Gr{\'e}goire Montavon},
  \bibinfo{person}{Sebastian Lapuschkin}, \bibinfo{person}{Alexander Binder},
  \bibinfo{person}{Wojciech Samek}, {and} \bibinfo{person}{Klaus-Robert
  M{\"u}ller}.} \bibinfo{year}{2017}\natexlab{}.
\newblock \showarticletitle{Explaining nonlinear classification decisions with
  deep taylor decomposition}.
\newblock \bibinfo{journal}{\emph{Pattern Recognition}}  \bibinfo{volume}{65}
  (\bibinfo{year}{2017}), \bibinfo{pages}{211--222}.
\newblock


\bibitem[\protect\citeauthoryear{Moraffah, Karami, Guo, Raglin, and
  Liu}{Moraffah et~al\mbox{.}}{2020}]%
        {moraffah2020causal}
\bibfield{author}{\bibinfo{person}{Raha Moraffah}, \bibinfo{person}{Mansooreh
  Karami}, \bibinfo{person}{Ruocheng Guo}, \bibinfo{person}{Adrienne Raglin},
  {and} \bibinfo{person}{Huan Liu}.} \bibinfo{year}{2020}\natexlab{}.
\newblock \showarticletitle{Causal interpretability for machine
  learning-problems, methods and evaluation}.
\newblock \bibinfo{journal}{\emph{ACM SIGKDD Explorations Newsletter}}
  \bibinfo{volume}{22}, \bibinfo{number}{1} (\bibinfo{year}{2020}),
  \bibinfo{pages}{18--33}.
\newblock


\bibitem[\protect\citeauthoryear{Murdoch, Singh, Kumbier, Abbasi-Asl, and
  Yu}{Murdoch et~al\mbox{.}}{2019}]%
        {murdoch2019interpretable}
\bibfield{author}{\bibinfo{person}{W~James Murdoch}, \bibinfo{person}{Chandan
  Singh}, \bibinfo{person}{Karl Kumbier}, \bibinfo{person}{Reza Abbasi-Asl},
  {and} \bibinfo{person}{Bin Yu}.} \bibinfo{year}{2019}\natexlab{}.
\newblock \showarticletitle{Interpretable machine learning: definitions,
  methods, and applications}.
\newblock \bibinfo{journal}{\emph{arXiv preprint arXiv:1901.04592}}
  (\bibinfo{year}{2019}).
\newblock


\bibitem[\protect\citeauthoryear{Nam, Kim, and Kim}{Nam et~al\mbox{.}}{2018}]%
        {nam2018text}
\bibfield{author}{\bibinfo{person}{Seonghyeon Nam}, \bibinfo{person}{Yunji
  Kim}, {and} \bibinfo{person}{Seon~Joo Kim}.} \bibinfo{year}{2018}\natexlab{}.
\newblock \showarticletitle{Text-adaptive generative adversarial networks:
  manipulating images with natural language}. In
  \bibinfo{booktitle}{\emph{Proceedings of the 32nd International Conference on
  Neural Information Processing Systems}}. \bibinfo{pages}{42--51}.
\newblock


\bibitem[\protect\citeauthoryear{Niu, Tang, Zhang, Lu, Hua, and Wen}{Niu
  et~al\mbox{.}}{2021}]%
        {niu2021counterfactual}
\bibfield{author}{\bibinfo{person}{Yulei Niu}, \bibinfo{person}{Kaihua Tang},
  \bibinfo{person}{Hanwang Zhang}, \bibinfo{person}{Zhiwu Lu},
  \bibinfo{person}{Xian-Sheng Hua}, {and} \bibinfo{person}{Ji-Rong Wen}.}
  \bibinfo{year}{2021}\natexlab{}.
\newblock \showarticletitle{Counterfactual vqa: A cause-effect look at language
  bias}. In \bibinfo{booktitle}{\emph{Proceedings of the IEEE/CVF Conference on
  Computer Vision and Pattern Recognition}}. \bibinfo{pages}{12700--12710}.
\newblock


\bibitem[\protect\citeauthoryear{N{\'o}brega and Marinho}{N{\'o}brega and
  Marinho}{2019}]%
        {nobrega2019towards}
\bibfield{author}{\bibinfo{person}{Caio N{\'o}brega} {and}
  \bibinfo{person}{Leandro Marinho}.} \bibinfo{year}{2019}\natexlab{}.
\newblock \showarticletitle{Towards explaining recommendations through local
  surrogate models}. In \bibinfo{booktitle}{\emph{Proceedings of the 34th
  ACM/SIGAPP Symposium on Applied Computing}}. \bibinfo{pages}{1671--1678}.
\newblock


\bibitem[\protect\citeauthoryear{Pan, Goyal, and Lee}{Pan
  et~al\mbox{.}}{2019}]%
        {pan2019question}
\bibfield{author}{\bibinfo{person}{Jingjing Pan}, \bibinfo{person}{Yash Goyal},
  {and} \bibinfo{person}{Stefan Lee}.} \bibinfo{year}{2019}\natexlab{}.
\newblock \showarticletitle{Question-conditioned counterfactual image
  generation for vqa}.
\newblock \bibinfo{journal}{\emph{arXiv preprint arXiv:1911.06352}}
  (\bibinfo{year}{2019}).
\newblock


\bibitem[\protect\citeauthoryear{Pearl}{Pearl}{2018}]%
        {pearl2018theoretical}
\bibfield{author}{\bibinfo{person}{Judea Pearl}.}
  \bibinfo{year}{2018}\natexlab{}.
\newblock \showarticletitle{Theoretical Impediments to Machine Learning With
  Seven Sparks from the Causal Revolution}. In
  \bibinfo{booktitle}{\emph{Proceedings of the Eleventh ACM International
  Conference on Web Search and Data Mining}}. \bibinfo{pages}{3--3}.
\newblock


\bibitem[\protect\citeauthoryear{Ribeiro, Singh, and Guestrin}{Ribeiro
  et~al\mbox{.}}{2016}]%
        {ribeiro2016should}
\bibfield{author}{\bibinfo{person}{Marco~Tulio Ribeiro},
  \bibinfo{person}{Sameer Singh}, {and} \bibinfo{person}{Carlos Guestrin}.}
  \bibinfo{year}{2016}\natexlab{}.
\newblock \showarticletitle{" Why should i trust you?" Explaining the
  predictions of any classifier}. In \bibinfo{booktitle}{\emph{Proceedings of
  the 22nd ACM SIGKDD international conference on knowledge discovery and data
  mining}}. \bibinfo{pages}{1135--1144}.
\newblock


\bibitem[\protect\citeauthoryear{Ronneberger, Fischer, and Brox}{Ronneberger
  et~al\mbox{.}}{2015}]%
        {ronneberger2015u}
\bibfield{author}{\bibinfo{person}{Olaf Ronneberger}, \bibinfo{person}{Philipp
  Fischer}, {and} \bibinfo{person}{Thomas Brox}.}
  \bibinfo{year}{2015}\natexlab{}.
\newblock \showarticletitle{U-net: Convolutional networks for biomedical image
  segmentation}. In \bibinfo{booktitle}{\emph{International Conference on
  Medical image computing and computer-assisted intervention}}. Springer,
  \bibinfo{pages}{234--241}.
\newblock


\bibitem[\protect\citeauthoryear{Salimans, Goodfellow, Zaremba, Cheung,
  Radford, and Chen}{Salimans et~al\mbox{.}}{2016}]%
        {salimans2016improved}
\bibfield{author}{\bibinfo{person}{Tim Salimans}, \bibinfo{person}{Ian
  Goodfellow}, \bibinfo{person}{Wojciech Zaremba}, \bibinfo{person}{Vicki
  Cheung}, \bibinfo{person}{Alec Radford}, {and} \bibinfo{person}{Xi Chen}.}
  \bibinfo{year}{2016}\natexlab{}.
\newblock \showarticletitle{Improved techniques for training gans}.
\newblock \bibinfo{journal}{\emph{Advances in neural information processing
  systems}}  \bibinfo{volume}{29} (\bibinfo{year}{2016}),
  \bibinfo{pages}{2234--2242}.
\newblock


\bibitem[\protect\citeauthoryear{Selvaraju, Cogswell, Das, Vedantam, Parikh,
  and Batra}{Selvaraju et~al\mbox{.}}{2017}]%
        {selvaraju2017grad}
\bibfield{author}{\bibinfo{person}{Ramprasaath~R Selvaraju},
  \bibinfo{person}{Michael Cogswell}, \bibinfo{person}{Abhishek Das},
  \bibinfo{person}{Ramakrishna Vedantam}, \bibinfo{person}{Devi Parikh}, {and}
  \bibinfo{person}{Dhruv Batra}.} \bibinfo{year}{2017}\natexlab{}.
\newblock \showarticletitle{Grad-cam: Visual explanations from deep networks
  via gradient-based localization}. In \bibinfo{booktitle}{\emph{Proceedings of
  the IEEE international conference on computer vision}}.
  \bibinfo{pages}{618--626}.
\newblock


\bibitem[\protect\citeauthoryear{Shwartz-Ziv and Tishby}{Shwartz-Ziv and
  Tishby}{2017}]%
        {shwartz2017opening}
\bibfield{author}{\bibinfo{person}{Ravid Shwartz-Ziv} {and}
  \bibinfo{person}{Naftali Tishby}.} \bibinfo{year}{2017}\natexlab{}.
\newblock \showarticletitle{Opening the black box of deep neural networks via
  information}.
\newblock \bibinfo{journal}{\emph{arXiv preprint arXiv:1703.00810}}
  (\bibinfo{year}{2017}).
\newblock


\bibitem[\protect\citeauthoryear{Simonyan, Vedaldi, and Zisserman}{Simonyan
  et~al\mbox{.}}{2013}]%
        {simonyan2013deep}
\bibfield{author}{\bibinfo{person}{Karen Simonyan}, \bibinfo{person}{Andrea
  Vedaldi}, {and} \bibinfo{person}{Andrew Zisserman}.}
  \bibinfo{year}{2013}\natexlab{}.
\newblock \showarticletitle{Deep inside convolutional networks: Visualising
  image classification models and saliency maps}.
\newblock \bibinfo{journal}{\emph{arXiv preprint arXiv:1312.6034}}
  (\bibinfo{year}{2013}).
\newblock


\bibitem[\protect\citeauthoryear{Sokol and Flach}{Sokol and Flach}{2018}]%
        {sokol2018glass}
\bibfield{author}{\bibinfo{person}{Kacper Sokol} {and} \bibinfo{person}{Peter~A
  Flach}.} \bibinfo{year}{2018}\natexlab{}.
\newblock \showarticletitle{Glass-Box: Explaining AI Decisions With
  Counterfactual Statements Through Conversation With a Voice-enabled Virtual
  Assistant.}. In \bibinfo{booktitle}{\emph{IJCAI}}.
  \bibinfo{pages}{5868--5870}.
\newblock


\bibitem[\protect\citeauthoryear{Srivastava, Murali, Dubey, and
  Mukherjee}{Srivastava et~al\mbox{.}}{2019}]%
        {srivastava2019visual}
\bibfield{author}{\bibinfo{person}{Yash Srivastava}, \bibinfo{person}{Vaishnav
  Murali}, \bibinfo{person}{Shiv~Ram Dubey}, {and} \bibinfo{person}{Snehasis
  Mukherjee}.} \bibinfo{year}{2019}\natexlab{}.
\newblock \showarticletitle{Visual question answering using deep learning: A
  survey and performance analysis}.
\newblock \bibinfo{journal}{\emph{arXiv preprint arXiv:1909.01860}}
  (\bibinfo{year}{2019}).
\newblock


\bibitem[\protect\citeauthoryear{Teney, Abbasnedjad, and van~den Hengel}{Teney
  et~al\mbox{.}}{2020}]%
        {teney2020learning}
\bibfield{author}{\bibinfo{person}{Damien Teney}, \bibinfo{person}{Ehsan
  Abbasnedjad}, {and} \bibinfo{person}{Anton van~den Hengel}.}
  \bibinfo{year}{2020}\natexlab{}.
\newblock \showarticletitle{Learning what makes a difference from
  counterfactual examples and gradient supervision}. In
  \bibinfo{booktitle}{\emph{Computer Vision--ECCV 2020: 16th European
  Conference, Glasgow, UK, August 23--28, 2020, Proceedings, Part X 16}}.
  Springer, \bibinfo{pages}{580--599}.
\newblock


\bibitem[\protect\citeauthoryear{Tsang, Liu, Purushotham, Murali, and
  Liu}{Tsang et~al\mbox{.}}{2018}]%
        {tsang2018neural}
\bibfield{author}{\bibinfo{person}{Michael Tsang}, \bibinfo{person}{Hanpeng
  Liu}, \bibinfo{person}{Sanjay Purushotham}, \bibinfo{person}{Pavankumar
  Murali}, {and} \bibinfo{person}{Yan Liu}.} \bibinfo{year}{2018}\natexlab{}.
\newblock \showarticletitle{Neural interaction transparency (nit):
  Disentangling learned interactions for improved interpretability}.
\newblock \bibinfo{journal}{\emph{Advances in Neural Information Processing
  Systems}}  \bibinfo{volume}{31} (\bibinfo{year}{2018}),
  \bibinfo{pages}{5804--5813}.
\newblock


\bibitem[\protect\citeauthoryear{Verma, Dickerson, and Hines}{Verma
  et~al\mbox{.}}{2020}]%
        {verma2020counterfactual}
\bibfield{author}{\bibinfo{person}{Sahil Verma}, \bibinfo{person}{John
  Dickerson}, {and} \bibinfo{person}{Keegan Hines}.}
  \bibinfo{year}{2020}\natexlab{}.
\newblock \showarticletitle{Counterfactual explanations for machine learning: A
  review}.
\newblock \bibinfo{journal}{\emph{arXiv preprint arXiv:2010.10596}}
  (\bibinfo{year}{2020}).
\newblock


\bibitem[\protect\citeauthoryear{Wu, Teney, Wang, Shen, Dick, and van~den
  Hengel}{Wu et~al\mbox{.}}{2017}]%
        {wu2017visual}
\bibfield{author}{\bibinfo{person}{Qi Wu}, \bibinfo{person}{Damien Teney},
  \bibinfo{person}{Peng Wang}, \bibinfo{person}{Chunhua Shen},
  \bibinfo{person}{Anthony Dick}, {and} \bibinfo{person}{Anton van~den
  Hengel}.} \bibinfo{year}{2017}\natexlab{}.
\newblock \showarticletitle{Visual question answering: A survey of methods and
  datasets}.
\newblock \bibinfo{journal}{\emph{Computer Vision and Image Understanding}}
  \bibinfo{volume}{163} (\bibinfo{year}{2017}), \bibinfo{pages}{21--40}.
\newblock


\bibitem[\protect\citeauthoryear{Yang, He, Gao, Deng, and Smola}{Yang
  et~al\mbox{.}}{2016}]%
        {yang2016stacked}
\bibfield{author}{\bibinfo{person}{Zichao Yang}, \bibinfo{person}{Xiaodong He},
  \bibinfo{person}{Jianfeng Gao}, \bibinfo{person}{Li Deng}, {and}
  \bibinfo{person}{Alex Smola}.} \bibinfo{year}{2016}\natexlab{}.
\newblock \showarticletitle{Stacked attention networks for image question
  answering}. In \bibinfo{booktitle}{\emph{Proceedings of the IEEE conference
  on computer vision and pattern recognition}}. \bibinfo{pages}{21--29}.
\newblock


\bibitem[\protect\citeauthoryear{Zhang, Goyal, Summers-Stay, Batra, and
  Parikh}{Zhang et~al\mbox{.}}{2016}]%
        {zhang2016yin}
\bibfield{author}{\bibinfo{person}{Peng Zhang}, \bibinfo{person}{Yash Goyal},
  \bibinfo{person}{Douglas Summers-Stay}, \bibinfo{person}{Dhruv Batra}, {and}
  \bibinfo{person}{Devi Parikh}.} \bibinfo{year}{2016}\natexlab{}.
\newblock \showarticletitle{Yin and yang: Balancing and answering binary visual
  questions}. In \bibinfo{booktitle}{\emph{Proceedings of the IEEE Conference
  on Computer Vision and Pattern Recognition}}. \bibinfo{pages}{5014--5022}.
\newblock


\bibitem[\protect\citeauthoryear{Zhang, Wang, Wu, Zhou, and Zhu}{Zhang
  et~al\mbox{.}}{2020}]%
        {zhang2020interpretable}
\bibfield{author}{\bibinfo{person}{Quanshi Zhang}, \bibinfo{person}{Xin Wang},
  \bibinfo{person}{Ying~Nian Wu}, \bibinfo{person}{Huilin Zhou}, {and}
  \bibinfo{person}{Song-Chun Zhu}.} \bibinfo{year}{2020}\natexlab{}.
\newblock \showarticletitle{Interpretable CNNs for object classification}.
\newblock \bibinfo{journal}{\emph{IEEE transactions on pattern analysis and
  machine intelligence}} \bibinfo{volume}{43}, \bibinfo{number}{10}
  (\bibinfo{year}{2020}), \bibinfo{pages}{3416--3431}.
\newblock


\bibitem[\protect\citeauthoryear{Zhang and Zhu}{Zhang and Zhu}{2018}]%
        {zhang2018visual}
\bibfield{author}{\bibinfo{person}{Quan-shi Zhang} {and}
  \bibinfo{person}{Song-chun Zhu}.} \bibinfo{year}{2018}\natexlab{}.
\newblock \showarticletitle{Visual interpretability for deep learning: a
  survey}.
\newblock \bibinfo{journal}{\emph{Frontiers of Information Technology \&
  Electronic Engineering}} \bibinfo{volume}{19}, \bibinfo{number}{1}
  (\bibinfo{year}{2018}), \bibinfo{pages}{27--39}.
\newblock


\bibitem[\protect\citeauthoryear{Zhang, Niebles, and Soto}{Zhang
  et~al\mbox{.}}{2019}]%
        {zhang2019interpretable}
\bibfield{author}{\bibinfo{person}{Yundong Zhang}, \bibinfo{person}{Juan~Carlos
  Niebles}, {and} \bibinfo{person}{Alvaro Soto}.}
  \bibinfo{year}{2019}\natexlab{}.
\newblock \showarticletitle{Interpretable visual question answering by visual
  grounding from attention supervision mining}. In
  \bibinfo{booktitle}{\emph{2019 ieee winter conference on applications of
  computer vision (wacv)}}. IEEE, \bibinfo{pages}{349--357}.
\newblock


\bibitem[\protect\citeauthoryear{Zhu, Mogadala, and Klakow}{Zhu
  et~al\mbox{.}}{2021}]%
        {zhu2021image}
\bibfield{author}{\bibinfo{person}{Dawei Zhu}, \bibinfo{person}{Aditya
  Mogadala}, {and} \bibinfo{person}{Dietrich Klakow}.}
  \bibinfo{year}{2021}\natexlab{}.
\newblock \showarticletitle{Image manipulation with natural language using
  Two-sided Attentive Conditional Generative Adversarial Network}.
\newblock \bibinfo{journal}{\emph{Neural Networks}}  \bibinfo{volume}{136}
  (\bibinfo{year}{2021}), \bibinfo{pages}{207--217}.
\newblock


\bibitem[\protect\citeauthoryear{Zhu, Mao, Liu, Zhang, Wang, and Zhang}{Zhu
  et~al\mbox{.}}{2020}]%
        {zhu2020overcoming}
\bibfield{author}{\bibinfo{person}{Xi Zhu}, \bibinfo{person}{Zhendong Mao},
  \bibinfo{person}{Chunxiao Liu}, \bibinfo{person}{Peng Zhang},
  \bibinfo{person}{Bin Wang}, {and} \bibinfo{person}{Yongdong Zhang}.}
  \bibinfo{year}{2020}\natexlab{}.
\newblock \showarticletitle{Overcoming language priors with self-supervised
  learning for visual question answering}.
\newblock \bibinfo{journal}{\emph{arXiv preprint arXiv:2012.11528}}
  (\bibinfo{year}{2020}).
\newblock


\end{thebibliography}

\end{document}